\documentclass[hidelinks, 12pt]{article}
\usepackage{moreverb,url}
\usepackage[colorlinks,bookmarksopen,bookmarksnumbered,citecolor=red,urlcolor=red]{hyperref}
\usepackage[utf8]{inputenc}
\usepackage{amsmath}
\usepackage{url, float, graphicx, algorithm, algorithmic, placeins}
\usepackage{color, xcolor}
\usepackage{siunitx}
\usepackage[small, bf]{caption}

\newcommand{\reals}{{\mbox{\bf R}}}

\newcommand{\symm}{{\mbox{\bf S}}}  % symmetric matrices

\newcommand{\conv}{\mathop{\bf conv}}

\newcommand{\eg}{{\it e.g.}}
\newcommand{\ie}{{\it i.e.}}

\newcommand{\BEAS}{\begin{eqnarray*}}
\newcommand{\EEAS}{\end{eqnarray*}}
\newcommand{\BEA}{\begin{eqnarray}}
\newcommand{\EEA}{\end{eqnarray}}
\newcommand{\BEQ}{\begin{equation}}
\newcommand{\EEQ}{\end{equation}}
\newcommand{\BIT}{\begin{itemize}}
\newcommand{\EIT}{\end{itemize}}

\usepackage{amsthm}

\usepackage{fullpage}

\newcommand{\normsq}[1]{\left\|{#1}\right\|_2^2}

\newcommand{\ellipse}{\mathcal{E}}
\newcommand{\eps}{\varepsilon}

\newcommand{\xgoal}{x^\mathrm{g}}

\title{Reciprocal Multi-Robot Collision Avoidance with Asymmetric State Uncertainty}

\author{Kunal Shah\thanks{These authors contributed equally.} \\ {\small \texttt{k2shah@stanford.edu}}\and Guillermo Angeris\footnotemark[1] \\{\small \texttt{angeris@stanford.edu}} \and Mac Schwager \\{\small \texttt{schwager@stanford.edu}} }
\date{May 2021}

\begin{document}
\maketitle
\begin{abstract}
%We present a fully distributed collision avoidance algorithm based on convex
%optimization for a team of mobile robots. This method addresses the practical 
%case in which agents sense each other via measurements from noisy on-board 
%sensors with no inter-agent communication. Under some mild conditions, we 
%provide guarantees on mutual collision avoidance for a broad 
%class of policies including the one presented. Additionally, we provide 
%numerical examples of computational performance and show that, in both 2D and 
%3D simulations, all agents avoid each other and reach their desired goals in 
%spite of their uncertainty about the locations of other agents.

We present a general decentralized formulation for a large class of collision
avoidance methods and show that all collision avoidance methods of this form
are guaranteed to be collision free. This class includes several existing
algorithms in the literature as special cases. We then present a particular
instance of this collision avoidance method, CARP (Collision Avoidance by
Reciprocal Projections), that is effective even when the estimates of other agents' positions
and velocities are noisy. The method's main computational step
involves the solution of a small convex optimization problem,
which can be quickly solved in practice, even on embedded platforms,
making it practical to use on computationally-constrained robots such as quadrotors. This method can be extended to find smooth polynomial trajectories for higher dynamic systems such at quadrotors. 
We demonstrate this algorithm's performance in simulations and on a team of physical
quadrotors.  Our method finds optimal projections in a median time of $17.12 \si{\ms}$ for $285$ instances of $100$ randomly generated obstacles, and produces safe polynomial trajectories at over $60 \si{hz}$ on-board quadrotors. Our paper is accompanied by an open source Julia implementation and ROS package.
\end{abstract}

%title
% sage_latex_guidelines.tex V1.20, 14 January 2017
%document sections
%intro
%!TEX root = ../root.tex
\section{Introduction}
Reliable collision avoidance is quickly becoming a mainstay requirement of any
scalable mobile robotics system. As robots continue to be deployed around
humans, assurances of safety become more critical, especially in high traffic
areas such as factory floors and hospital corridors. We define a class of
distributed collision avoidance methods, known as the reciprocally safe methods,
which we prove are guaranteed to be collision
free by construction. This class contains a number of well-known, published algorithms, providing
an alternative proof of collision avoidance.
We then present a special case of this class that allows a group of robots to avoid colliding with one another, even when each robot has its own (potentially noisy) estimates of other robots' states, as is common with noisy on-board sensors.
% This method allows each robot to have its own estimate of the relative positions of
% other robots, which may be inconsistent with other robots' estimates, and it 
The method additionally requires no explicit communication among the robots,
nor does it require these estimates to be consistent with those of other robots.  More specifically, we assume that each robot keeps
an uncertainty set (\eg, unions and intersections of ellipsoids) that contain the possible future locations of other robots.
% We assume each robot knows its own position
% exactly and updates its estimates of the other robots via noisy on-board sensors
% such as a camera or LIDAR.

The resulting policy is distributed in the sense that each robot \emph{only} requires an
estimate of the relative positions of the other robots. In other words, robots
do not need to communicate their positions to one another to coordinate their actions. Each robot uses its own position estimates for the other robots to find a
safe-reachable set for itself, which is characterized by a generalized Voronoi cell.
Our algorithm then computes a projection onto this safe-reachable set, which we show reduces to an
efficiently solvable convex optimization problem. We then extend the projection method to find smooth polynomial trajectories, instead of just a single point, which lie entirely inside the agents' safe-reachable set.
Our method is amenable to fast convex optimization solvers.  Using our method, a quadrotor maneuvering in a 3D space with 100 other quadrotos can compute its control action in approximately $17\si{\ms}$, including setup and solution time. Because the resulting method is reciprocally safe, we have the immediate implication that, if each robot uses this policy, mutual collision avoidance is
guaranteed.

 \begin{figure}[t]
    \centering
    \includegraphics[width=.48\textwidth]{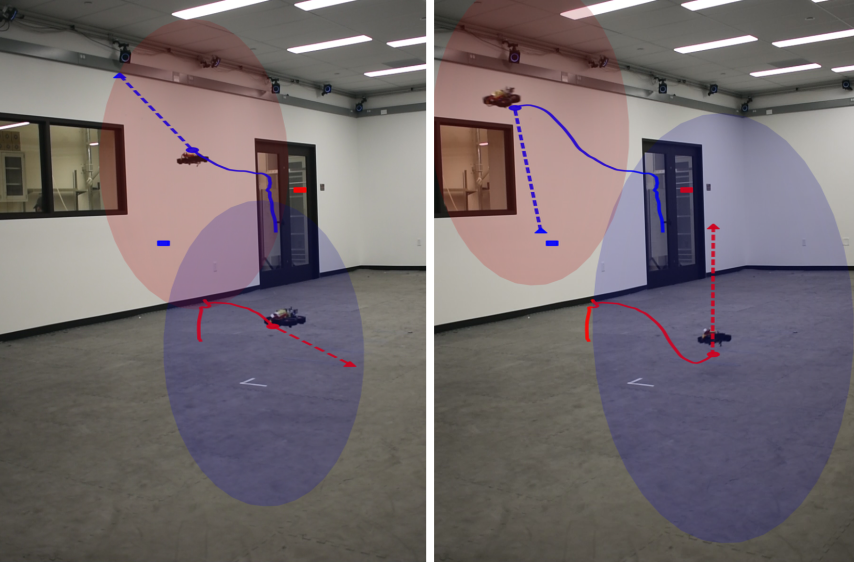} 
    \caption{Two quadrotors executing a reciprocal collision avoidance maneuverer each agent (red, blue) maintains an ellipsoidal estimate of the other's position. The agents move towards their goal point (square) projecting (triangle) it into their safe-reachable set. The solid curve shows the trajectory history and the dotted curve shows the path to the projected point.}
    \label{fig::hero}
\end{figure}

\paragraph{Summary.} This paper is organized as follows. The remainder of this section discusses
related work. The following section defines the class of
`reciprocally safe methods', proves that reciprocal collision avoidance is guaranteed
for any method in this class, and then discusses some basic, but useful, extensions
to this proof.
The following section presents a particular instance of a reciprocally
safe method, called `CARP,' short for Collision Avoidance by Reciprocal Projections. This method is specifically constructed
to guarantee safety, even in the presence of noisy sensor data,
while also being fast to execute on-board, as the resulting trajectory
can be computed by solving a small convex optimization problem. The following
section shows some benchmarks of the algorithm on synthetic data,
a large team of simulated robots in both 2D and 3D, and, finally, on a team of quadrotors
with noisy on-board data. We conclude with some thoughts and possible extensions. This paper extends the previous conference submission~\cite{carp1} by adding a smooth polynomial trajectory generation method via the reciprocal projection framework as well as hardware demonstrations on a quadrotor platform. It also simplifies and extends the proof and the constructions given in the original paper.
% The following section formulates the mutual avoidance problem and
% gives the necessary mathematical background on generalized Voronoi partitioning.
% Section~\ref{avoid} describes the collision-avoidance algorithm and provides a
% collision avoidance guarantee.  Section~\ref{cvxProj} formalizes the projection
% problem and describes the resulting convex optimization for ellipsoidal
% uncertainties in detail along with a method for smooth and safe trajectories quadrotor agents. Finally, section~\ref{results} shows our method's
% computational performance. Additionally, we show in 2D and 3D simulations and hardware demonstrations (figure \ref{fig::hero}) that
% all agents avoid each other and navigate to their goal locations despite their
% positional uncertainty of other agents.

\subsection{Related work}
The most closely related methods for fully distributed collision
avoidance in the literature are the velocity obstacle (VO) methods,
which can be used for a variety of collision avoidance strategies. 
These 
methods work by extrapolating the next position of an obstacle
using its current position and velocity. One of the most common tools used for
mutual collision avoidance is the Reciprocal Velocity Obstacles
(RVO)~\cite{rvo:2008,rvo_n:2011,ocra:2016} method in which each agent
solves a linear program to find its next step. The Buffered Voronoi
Cell (BVC) method of~\cite{bvc:2017} provides similar avoidance
guarantees, but does not require the ego agent to know other agents'
velocities, which can be difficult to estimate accurately. The BVC
algorithm opts instead for defining a given distance margin to compute
safe paths. BVC methods have been coupled with other decentralized
path planing tools~\cite{heirRobust:2019} in order to successfully
navigate more cluttered environments, but require that the other
agents' positions are known exactly.

\paragraph{Uncertainty.} While both VO and BVC methods scale very well to many (more than 100)
agents, they also require perfect state information of other agents'
positions (BVC), or positions and velocities (RVO). In many practical
cases, high accuracy state information, especially velocity, may not
be accessible as agents are estimating the position of the same
objects they are trying to avoid. Extensions to VO that account for
uncertainty have been studied under bounded~(\cite{boundedLoc:2012}) and
unbounded~(\cite{PRVO:2017}) localization uncertainties by utilizing
chance constraints. While these have been extended to decentralized
methods~(\cite{chanceCon:2019}), they assume constant velocity of the
obstacles at plan time. Unbounded localization estimates, modeled as Gaussian Mixture Models, were combined with RVO in~\cite{arul2021swarmcco}, and have been used effectively, but these methods only estimate the velocity at plan time and do not consider how the velocity of the other agents may change over the horizon. Combined Voronoi partitioning and estimation
methods have been studied for multi-agent path planning
tasks~\cite{transport:2014}, but still require communication to build
an estimate via consensus. The V-RVO method of~\cite{arul:2021} is the most similar to our method in that it augments the safe reachable set by adding RVO style constraints to the BVC as well as deflating the safe-reachable regions based on higher order dynamics. In contrast to these, our method does not require
any communication or velocity state information, nor does it require
the true position of the other agents. Instead, the algorithm uses
only an estimate of the current position of nearby agents and
their reachable set within some time horizon.

\paragraph{Comparison.} Our algorithm takes a desired trajectory or goal point (which
can come from any source, akin to~\cite{rvo:2008,bvc:2017}), and
returns a safe next step for an agent to take while accounting for
both the uncertainty and the physical extent of other agents.  The focus of 
this work is on fast, on-board refinement
rather than total path planning. More specifically, while the
algorithm presented could be used to reach a far away goal point, it
is likely more useful as a small-to-medium scale planner for reaching
waypoints in a larger, centrally or locally planned trajectory.

\paragraph{Higher order planners.} Similarly, single agent path planners such as A*~\cite{A-star} or
rapidly-exploring random trees (RRTs)~\cite{rrt} can be applied to the 
multi-agent case, but solution times grow rapidly due to the exploding 
size of the joint state space. For example, graph-search methods can be
partially decoupled~\cite{MSTAR:2011} to better scale for larger
multi-agent systems, but can explore the entire joint state-space in
the worst case. Fast Marching Tree (FMT) methods~\cite{fmt::2015} are
similar to RRTs in that they dynamically build a graph via
sampling. But, while FMT methods have better performance in higher
dimensional systems, they still require the paths to be centrally
calculated. A* can also
be used in dense environments for decentralized multi-agent planning when combined with barrier functions~\cite{barrier:2016},
but require the true position of all other agents. While all of 
these methods require global knowledge and large
searches over a discrete set, they can be used as waypoint
generators that feed into our method---for use in, \eg, cluttered
environments.

\paragraph{Nonconvex methods.} Optimization methods that use sequential convex programming
(SCP)~\cite{scp:2013,scpQuad:2012,scpSpace:2014} have also been studied for
multi-agent path planning; however, these algorithms are still centralized and
may exhibit slow convergence, making them unreliable for on-line planning. For
some systems, these methods can be partially decoupled~\cite{iscp:2015},
reducing computation time at the cost of potentially returning infeasible paths.
Our method, in comparison, is fully decentralized and produces an efficient convex
program for each agent. The solution of this program is a safe waypoint for the
agent, which, unlike SCP methods, requires no further refinement.
%Additionally, some end-to-end deep reinforcement learning
%methods~\cite{DeepRL:2017,DeepRL:2018} have been applied to this problem to
%produce controls as a function of state information but may not be certifiable
%for safety-critical applications. In contrast, our method provides a mutual
%collision avoidance guarantee if the group of agents in question all use the
%presented algorithm.

%prob
%!TEX root = ../root.tex
\section{Reciprocally safe methods}\label{probForm}
In this section we define a class of methods, called the reciprocally
safe methods, and show that any method in this class
is guaranteed to be collision-free, assuming the agents start out
in a collision-free configuration. We will then show that this proof
contains a number of results in the literature as special cases, and
give some simple extensions, such as ensuring a minimum separation.

\subsection{Method description}
We start with $n$ agents in an unbounded Euclidean space. For each agent $i=1, \dots, n$,
let $x_i(t) \in \reals^d$ denote its position, where $d$ is the dimension of the space, at time $t=1, \dots, T$. At each
time step $t$, each agent $i$ will have an estimate of possible future positions
of every other agent $j$, given by some set $\ellipse^i_j(t) \subseteq \reals^d$, with $i, j=1, \dots, n$, and $j \ne i$. (Note that this
set need not be bounded, nor closed for the proof.)
We will assume that $\ellipse^i_j(t)$ is \emph{consistent}; \ie,
that agent $j$'s position at time $t+1$ is always within the uncertainty set of agent $i$:
\[
x_j(t+1) \in \ellipse^i_j(t),
\]
for every agent $i, j=1, \dots, n$ with $i\ne j$ and every given time $t=1, \dots, T-1$.
%(We will see how to generalize this to the setting where $x_j(t+1)$ is instead in $\ellipse^i_j(t)$ with high
%probability in the next subsection.)

\paragraph{Reciprocally safe set.} A set $P_i(t) \subseteq \reals^d$ for agent $i$,
at time $t$, is \emph{reciprocally safe} if, for every other agent $j$ with $j \ne i$,
we have that
\begin{equation}\label{eq:r-safe}
P_i(t) \cap \ellipse^i_j(t) = \emptyset.
\end{equation}
One way of interpreting $P_i(t)$ is as a set of positions which are known to be safe to move to, given the position estimates
of every other agent. As before, we do not require this set be bounded or
even closed.

We will then say that the $n$ agents implement a \emph{reciprocally safe method} if there exists reciprocally safe sets
$P_i(t) \subseteq \reals^d$ such that
\[
x_i(t+1) \in P_i(t),
\]
if $P_i(t)$ is nonempty, and $x_i(t+1) = x_i(t)$, otherwise, for each $i=1, \dots, n$ and $t=1, \dots, T-1$. Written out,
we say that agents implement a reciprocally safe method whenever they move to a reciprocally safe position whenever such a position
is available, and do not move otherwise.
We will now show that any 
reciprocally safe method is collision-free, assuming the agents start from a collision-free configuration.

\paragraph{Proof.} The proof is almost by definition. Consider any two distinct agents $i, j=1, \dots, n$ with $i\ne j$ at some time 
$t=1, \dots, T$. We will show that $x_i(t+1)$ and $x_j(t+1)$ always have positive distance from each other, assuming that $x_i(t)$ and 
$x_j(t)$ are some positive distance apart. First, if $P_i(t)$ is nonempty, then,
by definition, we have
\[
x_i(t+1) \in P_i(t),
\]
but since the set $P_i(t)$ has empty intersection with the uncertainty set, by assumption,
\[
P_i(t) \cap \ellipse_j^i(t) = \emptyset,
\]
then,
\[
x_i(t+1) \not \in \ellipse_j^i(t),
\]
and, finally, since the uncertainty set contains the true position of agent $j$ at time $t+1$,
\[
x_j(t+1) \in \ellipse_j^i(t),
\]
so $x_i(t+1)$ and $x_j(t+1)$ must have positive separation.

On the other hand, if $P_i(t)$ is empty, and $P_j(t)$ is nonempty,
the proof follows similarly by replacing $i$ with $j$. Finally,
if both $P_i(t)$ and $P_j(t)$ are empty, then
\[
x_i(t+1) = x_i(t), \quad \text{and} \quad x_j(t+1) = x_j(t),
\]
so $x_i(t+1)$ and $x_j(t+1)$ have positive separation because $x_i(t)$ and $x_j(t)$
did, by assumption.

Because this is the case for any $i \ne j$ and every $t=1, \dots, T-1$, any reciprocally
safe method is collision free as any two agents always have positive separation,
assuming they have positive separation at $t=1$.

\subsection{Basic extensions}\label{sec:extensions}
We outline some immediate extensions to the proof above.

\paragraph{Positive margin.} In the previous proof, while there is a guarantee that the
agents will be positively separated, the separation could be arbitrarily small. On the
other hand, we can imagine that the agents are all required to have some amount
of margin that is bounded from below. We will denote the required margin by some set $B \subseteq \reals^d$,
which we view as a ``safe" region around the agent. Usually $B$ will be a ball in $d$ dimensions:
\[
B = \{y\mid \|y\|_2 \le \eps\},
\]
where $\eps > 0$ is the desired minimal distance from all other agents, but we may generally
take any set $B$ we wish.

To guarantee this, we can strengthen condition~\eqref{eq:r-safe} to
\[
(P_i(t) + B) \cap \ellipse_j^i(t) = \emptyset, 
\]
whenever the set $P_i(t)$ is nonempty. Here, we define
\[
P_i(t) + B = \{y + z \mid y \in P_i(t), ~ z \in B\},
\]
as the set sum, or Minkowski sum, of $P_i(t)$ and $B$. Another way of stating this is, if the set $P_i(t)$ is nonempty,
it must only contain points which have a margin of at least $B$ from the future estimated positions.
The resulting proof is similar to the base method and gives the stronger guarantee that
\[
x_j(t+1) \not \in x_i(t+1) + B,
\]
for any two agents $i \ne j$. In the special case that $B$ is an $\eps$-ball, this would imply that
agents $i$ and $j$ are at least $\eps$ distance apart.

\paragraph{Stopping set.} The proof above requires the condition that
$x_i(t+1) = x_i(t)$ if the set $P_i(t)$ is empty. This is, of course,
not always possible for realistic agents such as drones, which cannot stop immediately.
In this case, we can relax the condition $x_i(t+1) = x_i(t)$ to the following condition:
\[
x_i(t+1) \in x_i(t) + S_i = \{x_i(t) + y \mid y \in S_i\},
\]
where $S_i \subseteq \reals^d$ is agent $i$'s \emph{stopping set} (similar to the braking set used in \cite{arul:2021}) which is assumed to be
compact. We then strengthen the initial condition that $x_i(t)$ and $x_j(t)$ are positively separated for $i\ne j$, to
\[
(x_i(t) + S_i) \cap (x_j(t) + S_j) = \emptyset,
\]
and the reciprocal safety condition that $P_i(t)$ must satisfy when it is nonempty, to
\[
(P_i(t) + S_i) \cap \ellipse_j^i(t) = \emptyset,
\]
for each $i, j=1, \dots, n$ with $i\ne j$.
%where
%\[
%P_i(t) + S_i = \bigcup_{y \in S_i}P_i(t) + y.
%\]
Positive separation between $x_i(t+1) \in x_i(t) + S_i$
and $x_j(t+1) \in x_j(t) + S_j$ is then always guaranteed in this case by a similar proof, since
disjoint compact sets have positive separation. Note that this only guarantees one-step separation, from time
$t$ to time $t+1$.
We can then guarantee positive separation at all times
by requiring the additional condition that, if $P_i(t)$ and $P_i(t+1)$ are both empty, we have
\[
x_i(t+2) = x_i(t+1),
\]
\ie, agent $i$ only needs to `stop once.'

%\paragraph{Probabilistic constraints.} \blue{idk if we NEED this, like it's nice but if you take this probability over the entire horizon (infinite) the prob of collision is 1 right? This is kind of why i just ignored the unbounded probs. this aside is short so it's not that big if we leave it in. just my two cents} We can also weaken the consistency constraint on $\ellipse_j^i(t)$
%to a probabilistic constraint:
%\[
%x_j(t+1) \not\in \ellipse_j^i(t) ~~ \text{with probability} ~~ p_{ij},
%\]
%for each $i, j=1, \dots, n$ with $i\ne j$. A simple union-bound gives that
%\[
%\Prob(\text{no collision}) \ge 1 - \sum_{i\ne j} p_{ij}.
%\]
%In the case that $p_{ij}$ are independent events over each agent $i$, this can
%be strengthened to
%\[
%\Prob(\text{no collision}) \ge 1 - \sum_{j}\left(1-\prod_{i\ne j}(1-p_{ij})\right),
%\]
%and similarly over agent $j$. There are obvious extensions to more granular bounds
%when there is more information about the joint probabilities.

\subsection{Example methods}
There are several simple methods which satisfy the reciprocal safety property,
and are therefore guaranteed to be collision free. We describe some basic examples
in this section, which include some methods in the literature that are known to be
collision free by other proof techniques, showing that this class is general enough
to include a number of known results.

\paragraph{Trivial method.} Perhaps the simplest of all possible reciprocally safe
methods is the \emph{trivial method}, which, for all agents $i=1, \dots, n$, defines
$P_i(t)$ as
\[
P_i(t) = \emptyset, \quad  t =1, \dots, T-1,
\]
and sets the position estimates to be all of $\reals^d$ for any distinct agents $i, j=1, \dots, n$
with $i\ne j$,
\[
\ellipse_j^i(t) = \reals^d, \quad t=1, \dots, T-1.
\]
In other words, the agents' position estimates of each other are `maximally bad,' \ie, they
include all of $\reals^d$, and there is no safe position to move to, as $P_i(t) = \emptyset$
for each agent $i$ and time $t$. It is follows that the sets $P_i(t)$ satisfy
the reciprocal safety conditions~\eqref{eq:r-safe}, and that the agents must always have $x_i(t+1) = x_i(t)$.
Because of this, the agents never move from their starting locations,
which is easily seen to be collision free, when starting from a collision free configuration.

While this example is not useful in practice, it is a good initial exercise to check the conditions
necessary for a method to be reciprocally safe.

\paragraph{Buffered Voronoi cell.} Another reciprocally safe method is the
buffered Voronoi cell method of~\cite{bvc:2017}.
In this method, the future position estimate that agent $i$ has of agent $j$ at time $t$, is given by
\[
\ellipse_j^i(t) = \{y \in \reals^d \mid \|y - x_j(t)\|_2^2 + \eps \le \|y - x_i(t)\|_2^2\},
\]
where $x_j(t)$ is the (known) position of agent $j$ at time $t$, $x_i(t)$ is the position of agent $i$,
and $\eps > 0$ denotes some margin. In other words, the possible future
positions of agent $j$ are the set of points which are further from agent $i$ than from $j$
by at least $\eps$. (We take a simpler approach here than that of~\cite{bvc:2017} for the sake
of presentation, but the proof is nearly identical.)
%\blue{how different is it? i think this statements might rub people the wrong way if they don't read the other work and might be mad we didn't do our "due diligence" either we should remove this or do the OG formulation}
In this case, the set of `safe' locations for agent $i$ is defined similarly:
\[
P_i(t) = \bigcap_{j\ne i}\{y \in \reals^d \mid \|y - x_i(t)\|_2^2 + \eps \le \|y-x_j(t)\|_2^2\}.
\]
The reciprocal safety condition can be readily verified, since for distinct agents $i$
and $j$, and any point in the uncertainty region $y \in \ellipse_j^i(t)$ satisfies
\[
\|y - x_j(t)\|_2^2 + \eps \le \|y - x_i(t)\|_2^2,
\]
by definition. So, $y$ cannot be in the reciprocally safe set $P_i(t)$, since $y \in P_i(t)$
means that $y$ also satisfies,
\[
\|y - x_i(t)\|_2^2 + \eps \le \|y-x_j(t)\|_2^2 \le \|y - x_i(t)\|_2^2 - \eps,
\]
a contradiction.

Any method that agent $i$ uses to choose its next location within the set $P_i(t)$
is guaranteed to be collision free by the above proof. Additionally, because the set
$P_i(t)$ is a convex set as it is the intersection of a number of convex sets,
many convex optimization problems can easily include the constraint
that the agent's future position must lie in this set. Because convex problems are almost
always efficiently solvable, even with on-board computational constraints, finding feasible
points that best satisfy some requirement can often be done in real time.

\paragraph{Method refinement.} Very generally, we have the following additional result:
if a method is reciprocally safe, any refinement of such a method is also always safe. That is,
given a reciprocally safe method, where the agents $i\ne j$ have estimates $\ellipse^i_j(t)$
and reciprocally safe sets $P_i(t)$ at
each time $t$, we say a second method is a \emph{refinement}
of the first if it has estimates $\bar \ellipse^i_j(t)$ that satisfy $\bar \ellipse^i_j(t)\subseteq \ellipse^i_j(t)$, 
and reciprocally safe sets $\bar P_i(t)$ that satisfy $\bar P_i(t)\subseteq P_i(t)$.
This implies that if the original method is reciprocally safe,
\ie, satisfies~\eqref{eq:r-safe}, then a refinement is also reciprocally safe as it also satisfies~\eqref{eq:r-safe}, and is therefore 
collision free. More intuitively, this is can be restated as the fact that having better estimates, or a more restricted action space,
can never make an algorithm unsafe. This construction then immediately includes, for example, the safety results
of~\cite{arul:2021}, where the position estimates are the buffered Voronoi cells given in the previous example,
intersected with an additional reciprocal velocity obstacle (RVO) cone.

\section{General method}\label{avoid}
In this section, we present a reciprocally safe method that generalizes
the buffered Voronoi cells of~\cite{bvc:2017} to include measurement uncertainty
and higher order dynamics. The resulting method relies on the solution
of a small convex optimization problem at each time step that is unlikely
to have a closed form, but can still be solved with on-board systems
in under a millisecond with modern solvers.

To do this, we first introduce the idea of generalized Voronoi cells---a natural
way of extending Voronoi cells from collections of points to collections of sets.
In many cases, the resulting generalized Voronoi cells cannot be defined in terms of a finite
number of closed form inequalities, but, because these sets are always convex, we can write new
expressions that depend on a larger number of variables that do have closed forms.

\subsection{Generalized Voronoi cells}
Given a point $x \in \reals^d$ and a collection of $m$ sets $S_1, \dots, S_m \subseteq \reals^d$, we
will define the \emph{generalized Voronoi cell} of $x$ with respect to the family $S$ as
\begin{equation}\label{eq:generalized-voronoi}
V(x, S) = \{y \in \reals^d \mid \|y - x\|_2 \le \min_{j=1,\dots,m}\|y - S_j\|_2\}.
\end{equation}
Here, we have defined $\|y - S_j\|_2$ to be the distance-to-set function:
\[
\|y - S_j\|_2 = \inf_{z \in S_j} \|y - z\|_2,
\]
for $j=1, \dots, m$. (If $S_j$ is empty, we set the distance as $+\infty$, for convenience.)
We can view the points in $V(x, S)$ as the set of points which are closer to $x$ than to any
point in any one of the sets $S_j$. Note that we can write
\[
V(x, S) = \bigcap_{j=1}^m V(x, \{S_j\}).
\]
We will make use of this fact later in what follows.

\paragraph{Comparison to Voronoi cells.} The usual definition of the Voronoi cell
is, given a family of points $x_1, \dots, x_m \in \reals^d$, the cell generated for
the $i$th point is defined as the set of all points closer to $x_i$ than to any other
point $x_j$; \ie,
\[
\{y \in \reals^d \mid \|y - x_i\|_2 \le \min_{j=1, \dots, m} \|y - x_j\|_2 \}.
\]
This is the special case of~\eqref{eq:generalized-voronoi} where we take the generalized Voronoi
cell of $x_i$ with respect to $S$, and every set in the
family $S$ is a singleton; \ie, $S_j = \{x_j\}$ for each $j=1, \dots, m$.

\paragraph{Convexity.} Like the usual definition of a Voronoi cell, the generalized
Voronoi cell of $x$ with respect to $S$ is convex, even when the sets $S_j$ are not.
To see this, note that we can write $V$ as
\[
V(x, S) = \bigcap_{j=1}^m\bigcap_{z \in S_j} \{y \in \reals^d \mid \|y-x\|_2 \le \|y - z\|_2\},
\]
which is the intersection of a family of hyperplanes. (This follows by
squaring both sides of the inequality and cancelling the $\|y\|_2^2$ term.)
Because the intersection of convex sets is convex, and hyperplanes
are convex sets, then $V(x, S)$ is always convex.

%\paragraph{Collections of sets.} We will define the generalized Voronoi
%cell of $x$ with respect to a collection of $n$ sets $S_1, \dots, S_n \subseteq \reals^d$
%as the generalized Voronoi cell of $x$ with respect to the union of such
%sets, \ie,
%\[
%    V(x, S) = V\left(x, \bigcup_{i=1}^n S_i\right).
%\]
%Note that this matches the usual definition of a Voronoi cell in the case that
%the sets $S_i$ are singletons. We can interpret $V(x, S)$ as a Voronoi cell with
%respect to points with uncertainty: \eg, the exact position of point $i$, say
%$y_i \in \reals^d$, is not known, but the point is known to lie in some set
%$y_i \in S_i$. It is also not hard to show that $V(x, S)$ can be equivalently
%written as
%\begin{equation}\label{eq:voronoi-intersection}
%    V(x, S) = \bigcap_{i=1}^n V(x, S_i).
%\end{equation}

\subsection{Projective method}
We will now discuss a simple version of the method and its applications
to single-integrator dynamics. Later in this section, we will also
show how to extend this to more general dynamics
and more general sets.

\paragraph{Projection.} A common objective for an agent to optimize is its
distance to some desired goal. We will call this goal point $x_i^g \in
\reals^d$, for agent $i$. At every time step $t$, the optimization problem
for agent $i$ is then:
\[
    \begin{aligned}
        & \text{minimize} && \|y - x_i^g\|_2\\
        & \text{subject to} && y \in P_i(t).
    \end{aligned}
\]
The optimization variable here is $y \in \reals^d$, while the problem data are
the goal point $x_i^g \in \reals^d$ and the reciprocally safe set $P_i(t)$.
If a solution to the problem $y^\star$ exists, the agent takes a step towards
$y^\star$, otherwise, if there are no feasible points, then the agent stops in place.
(We will see extensions to the more general case where the agent has some stopping
distance later in this section.)

Such a method is often called a ``greedy'' method, as the agent attempts to get
as close as possible as it can to the goal position, while remaining safe. We
refer to this specific way of picking the next possible point as a `projective
method' since a solution $y^\star$ is often called the projection of $x_i^g$ onto
the set $P_i(t)$. We will show how to construct reciprocally safe sets $P_i(t)$
for all agents $i=1, \dots, n$, assuming that each agent has some (potentially
noisy) estimate of the future locations of all other agents.

%We note that problem~\eqref{eq:main} is, in theory, a convex optimization
%problem, since it has a convex objective function and convex constraints.
%While it is likely efficiently solvable in practice, it is not clear how to
%encode the constraint $y \in P_i(t)$ in terms of inequalities of well known
%convex functions, except in very special cases. For example, in the case that
%the sets $\ellipse_i^j(t)$ are singletons, $P_i(t)$ is simply the Voronoi
%cell generated by the $\ellipse^i_j(t)$ for each $j$. The resulting set
%$P_i(t)$ is therefore a polyhedron and can be specified by a finite number of
%linear inequality constraints.

%We will show a more general method where the sets $\ellipse_i^j(t)$ are allowed to be
%either intersections or unions of convex sets such as polyhedra and ellipsoids. The
%method can likely be applied more generally to other sets and we encourage readers
%to do so. 

\paragraph{Safe estimates.} 
To simplify notation, we will write the algorithm for a single agent $i$ at a given time step
$t$. Because of this, we write $x$ for $x_i(t)$, which is
the agent's current position at time $t$, and $\ellipse_j$ for $\ellipse_j^i(t)$, which are the
estimates agent $i$ has of agent $j$ at the next time step, $t+1$.

Using this notation,
we can view the set $V(x, \ellipse)$
(\ie, the set of points which are closer to $x$ than they are to
any $\ellipse_j$) as a set of positions which agent $i$ is guaranteed
to reach before any agent $j$. More specifically:
\begin{equation}\label{eq:safe-set}
    V(x, \ellipse) \cap \ellipse_j = \emptyset,
\end{equation}
for each $j\ne i$. We also note that this set is an
overly-conservative set as there are many sets that also
satisfy~\eqref{eq:safe-set} that strictly contain $V(x,
\ellipse)$.

%Since this is true for each agent $j$, a simple choice of a reciprocally safe
%set is the intersection of all such sets, \ie,
%\[
%    P_i(t) = \bigcap_{j\ne i} V(x_i(t), \ellipse^i_j(t)).
%\]
A natural choice for a reciprocally safe set for agent $i$,
is then
\[
P_i(t) = V(x, \ellipse),
\]
since we know that any choice of $y \in P_i(t)$ is
guaranteed to be safe, by construction. This implies that, if all agents $i$ choose points
within $P_i(t)$ (and simply stop if the set is empty) then the method is
guaranteed to be collision free. The resulting optimization problem is:
\begin{equation}\label{eq:proj}
    \begin{aligned}
        & \text{minimize} && \|y - x_i^g\|_2\\
        & \text{subject to} && y \in V(x, \ellipse),
    \end{aligned}
\end{equation}
with variable $y \in \reals^d$. Note that, since $P_i(t)$ is a generalized Voronoi cell,
then $P_i(t)$ is convex. We will use this fact to give an efficient method for optimizing a goal function,
when the sets $\ellipse_j$ are the intersections and unions of well-known convex sets.

\subsection{Projecting onto generalized Voronoi cells}
%In this subsection, we describe a solution to the problem of efficiently projecting
%a point onto a generalized Voronoi cell.

%First, we construct a program which is equivalent to finding a projection into a
%convex set of the form~\eqref{eq:proj}, but may not be easy to solve as its
%constraint is not representable in any standard form. We then use Lagrange
%duality to construct a convex constraint that is at least as strict as the
%original and use strong duality to show that the two problems are equivalent.
%Finally, we provide a conic problem for
%the case of ellipsoid generated Voronoi regions with the constraint explicitly
%parametrized by the ellipsoid parameters $(\mu, \Sigma)$.  We also show that,
%for the ellipsoidal case, the constraint is represented by a sum of
%quadratic-over-linear terms, implying that the convex program is a second order
%cone program (SOCP) and can therefore be solved quickly by embedded solvers.
%As in~\cite{bvc:2017}, in order to execute the collision-avoidance strategy,
%each agent must project its goal point onto its safe-reachable set as defined
%in~\eqref{eq:main}. This problem is always convex since the Voronoi region is
%generated by an (arbitrary) intersection of hyperplanes~\cite[\S2.3.1]{cvxbook},
%which always results in a convex set.

We will show how to efficiently solve~\eqref{eq:proj} by first reducing it to
a problem over several constraints, each of which are simpler than the original. We then show how these constraints can be
reduced to a number of inequalities which are easily compiled down to well-known conic
constraints, when the uncertainty sets $\ellipse_j$ are unions and intersections of ellipsoids and
polygons. Similarly to the previous subsection, we will only consider agent $i$'s position at time $t$, denoted simply
as $x$ and the uncertain estimates of the future positions of the other agents as $\ellipse_j$ for $j=1, \dots, n$ with
$j \ne i$.

\paragraph{A basic reduction.} We first note that, given a problem of
the form of~\eqref{eq:proj}, we can write the equivalent problem:
\[
    \begin{aligned}
        & \text{minimize} && \|y - x_i^g\|_2\\
        & \text{subject to} && y \in V(x, \{\ellipse_j\}), \quad j \ne i.
    \end{aligned}
\]
In other words, we have `split' the single constraint $y \in V(x, \ellipse)$
to $n$ constraints given by $y \in V(x, \{\ellipse_j\})$ for each $j$. This follows
from the fact that
\[
V(x, \ellipse) = \bigcap_{j\ne i} V(x, \{\ellipse_j\}),
\]
which is readily verified from~\eqref{eq:generalized-voronoi}.
From this, it suffices to show how to write the constraint $y \in V(x, \{\ellipse_j\})$
for a single set $\ellipse_j$. We will write $\ellipse_1$ for the set in question,
with the understanding that $\ellipse_1$ stands for any `anonymous' set.

\paragraph{Convex sets.} In general, the set $\ellipse_1$ is defined by
\begin{equation}\label{eq:function}
    \ellipse_1 = \{z \in \reals^d \mid f(z) \le 0\},
\end{equation}
where $f: \reals^d \to \reals^r$ is a convex function. For example, in the case that $\ellipse_1$
is an ellipse defined by $(\mu, \Sigma)$ where $\mu \in \reals^d$ and $\Sigma \in \symm_{++}^d$
is positive definite, we have that
\[
f(z) = \frac12 z^T\Sigma^{-1}z + \mu^Tz -1.
\]
On the other hand, if $\ellipse_1$ is a polyhedron, it is defined by
a number of affine inequalities; \ie,
\[
f(z) = Az - b,
\]
where $A \in \reals^{r \times d}$ and $b \in \reals^r$. There are a number
of other possible functions, such as indicator sets among many others, but
we focus on these two cases as the most common types of sets.

\paragraph{Constraint rewriting.} Given any set $\ellipse_1$ defined by
a function $f$, as in~\eqref{eq:function}, the corresponding constraint is:
\[
y \in V(x, \{\ellipse_1\}).
\]
From~\eqref{eq:generalized-voronoi}, this is true, if, and only if, $y$ also satisfies
\[
\|y - x\|_2^2 \le \inf_{z \in \ellipse_1}\|y - z\|_2^2.
\]
We will rewrite this as
\begin{equation}\label{eq:inequality}
\|x\|_2^2 - 2y^Tx\le \inf_{f(z) \le 0}\left(\|z\|_2^2 - 2y^Tz\right),
\end{equation}
by expanding the squared norm on both sides, cancelling like terms, and using the definition of
$\ellipse_1$.
In general, it is unlikely that there is a closed form solution for the right hand side
of the inequality, even in the special cases where $\ellipse_1$ is an ellipse or a polyhedral
set. To get around this, we will use a duality trick, introduced originally in~\cite{carp1},
to rewrite the right hand side as an unconstrained infimum. This new infimum has a simple
analytical solution in the important cases where $f$ is an affine function (when $\ellipse_1$ is a
polyhedron) and when $f$ is a convex quadratic (when $\ellipse_1$ is an ellipsoid). There are likely
more applications of this method to more complicated functions $f$, but we focus on these two
important cases. We encourage readers to apply this method to other functions $f$ which may
be useful in practice.

\paragraph{Weak duality.} One simple approach to finding a reasonable replacement for~\eqref{eq:inequality}
is to find an approximation of the right hand side that is concave and reasonably tight.
One standard approach is by \emph{Lagrange duality}~\cite{cvxbook},
where the `hard constraint' $f(z) \le 0$ is relaxed to a linear penalty term with some weights $\lambda \in \reals_+^n$, \ie,
inequality~\eqref{eq:inequality} is replaced with
\begin{equation}\label{eq:restriction}
\|x\|_2^2 - 2y^Tx\le \inf_z\left(\|z\|_2^2 - 2y^Tz + \lambda^Tf(z)\right).
\end{equation}
This new infimum is unconstrained and sometimes, but not always, admits closed form solutions when the original
does not. (In many cases, the closed form solutions, when they exist, are well-known.)
For convenience, we will define the \emph{dual function} $g$ as
\[
g(y, \lambda) = \inf_z\left(\|z\|_2^2 - 2y^Tz + \lambda^Tf(z)\right),
\]
and note that, for any $\lambda \ge 0$ and $y$ we have that
\begin{equation}\label{eq:weak-duality}
    g(y, \lambda) \le \inf_{f(z) \le 0}\left(\|z\|_2^2 - 2y^Tz\right),
\end{equation}
which is known as \emph{weak duality}~\cite{cvxbook}.
Replacing inequality~\eqref{eq:inequality} with inequality~\eqref{eq:restriction} is often called a \emph{restriction}:
if $y$ is feasible for some $\lambda \ge 0$ in~\eqref{eq:restriction} then $y$ is also feasible for~\eqref{eq:inequality}.
In other words, the new constraint is at least as tight as the original. An important fact of the dual function $g$ is that it is concave in its arguments,
because $g(y, \lambda)$ is an infimum over a family of functions that are affine in $y$ and $\lambda$. This implies that~\eqref{eq:restriction}
is a convex inequality constraint.

\paragraph{Strong duality.} Due to strong duality, the new constraint is, in fact, equivalent to the original. That is,
for every $y$, there exists some $\lambda \ge 0$ such that inequality~\eqref{eq:weak-duality} holds at equality:
\[
g(y, \lambda) = \inf_{f(z) \le 0}\left(\|z\|_2^2 - 2y^Tz\right).
\]
Because of this, if $y$ is feasible for~\eqref{eq:inequality} then there exists some $\lambda$ such that $y$ is also feasible
for~\eqref{eq:restriction}, and vice versa. In other words, replacing inequality~\eqref{eq:inequality} with~\eqref{eq:restriction}
and solving this new problem (with $\lambda \ge 0$ as an additional variable) gives two equivalent problems in that a feasible $y$
for the first corresponds to a feasible pair $(y, \lambda)$, with the same objective value, for the second.

\paragraph{Dual functions for known sets.} Given that both constraints are equivalent, the last remaining point is to write down the dual functions for some 
known sets. We will make use of the fact that the minimizer of a convex quadratic is
\begin{equation}\label{eq:quad-min}
\inf_z \left(z^TPz + 2q^Tz\right) = -q^TP^{-1}q,
\end{equation}
for any positive definite matrix $P \in \symm_{++}^d$ and vector $q \in \reals^d$. This follows from 
an application of the first order optimality conditions.

\paragraph{Polyhedral uncertainty.} In the case where $\ellipse_1$ is a polyhedron, we have that
\[
f(z) = Az - b
\]
for some $A \in \reals^{r\times d}$ and $b \in \reals^r$. In this case, the dual function is
\[
g(y, \lambda) = \inf_z\left(\|z\|_2^2 - 2y^Tz + 2\lambda^T(Az - b)\right),
\]
where we have replaced $\lambda$ with $2\lambda$ for convenience. Using~\eqref{eq:quad-min}
and the fact that $\|z\|_2^2 = z^TIz$, we get:
\[
g(y, \lambda) = -\|A^T\lambda - y\|_2^2 - 2\lambda^Tb,
\]
as required.

\paragraph{Ellipsoidal uncertainty.} The case where $\ellipse_1$ is an ellipsoid with parameters $\mu \in \reals^d$ and $\Sigma \in \symm_{++}^d$
is rather similar. In this case, the function $f$ is a quadratic and the dual function is
\[
g(y, \lambda) = \inf_z\left(\|z\|_2^2 - 2y^Tz + \lambda(z^T\Sigma z + 2\mu^Tz - 1)\right).
\]
Replacing, again, $\lambda$ with $2\lambda$ for convenience. Collecting the quadratic, linear, and constant
terms and applying~\eqref{eq:quad-min}, we have:
\[
g(y, \lambda) = -(\lambda \mu - y)^T(I + \lambda\Sigma)^{-1}(\lambda\mu - y) - \lambda.
\]
This function is concave, and its corresponding constraint is easily representable as a semidefinite constraint.
On the other hand, most embedded solvers, often due to space and time restrictions, do not support semidefinite constraints. We can turn
this into a constraint that is representable as a small family of second-order cone constraints, which are often more efficient
in practice and can be handled by embedded solvers such as ECOS~\cite{ecos:2013}.

Since $\Sigma$ is positive definite, it has an eigendecomposition given by
\[
\Sigma = VDV^T,
\]
where $D \in \reals^{d\times d}$ is a diagonal matrix and $V \in \reals^{d\times d}$ is an orthogonal matrix such that $V^TV = VV^T = I$.
Using this, we can write:
\[
(I + \lambda\Sigma)^{-1} = (V(I + \lambda D)V^T)^{-1} = V(I + \lambda D)^{-1}V^T.
\]
This implies that
\[
g(y, \lambda) = -(V^T(\lambda \mu - y))^T(I + \lambda D)^{-1}(V^T(\lambda \mu - y)) - \lambda.
\]
Because the matrix $I + \lambda D$ is diagonal, then:
\begin{equation}\label{eq:ellipse}
g(y, \lambda) = -\sum_{i=1}^d \frac{(v_i^T(\lambda \mu - y))^2}{1 + \lambda D_{ii}} - \lambda,
\end{equation}
where $v_i$ is the $i$th column of $V$.
This expression is a sum of quadratic-over-linear terms, which are easily representable as
second-order cone constraints~\cite{lobo:1998} and are supported in most embedded solvers.

\paragraph{Unions and intersections of sets.} If $\ellipse_1$ is the union of a number
of ellipsoids or polyhedra, then, as before, we can simply split $\ellipse_1$ into its
individual components and add each as a constraint. On the other hand, $\ellipse_1$ can
also be the \emph{intersection} of ellipsoids and polyhedra; \ie, $\ellipse_1$ is the intersection
of the polyhedra specified by $A \in \reals^{r \times d}$ and $b \in \reals^{r}$
(as the intersection of polyhedra results in another polyhedron) and the ellipsoids
given by $(\mu_i, \Sigma_i)$ for $i=1, \dots, s$.

The derivation here is again nearly identical to the previous. We will write $\lambda_0 \in \reals_+^r$
as the Lagrange multiplier for the polyhedral constraint, while we write $\lambda_i \ge 0$ for $i=1, \dots, s$
for each of the ellipsoidal constraints. The dual function is then:
\[
g(y, \lambda) = -h\left(A^T\lambda_0 + \sum_{i=1}^s \lambda_i \mu_i - b, I + \sum_{i=1}^s \lambda_i\Sigma_i\right),
\]
where the function $h: \reals^d \times \symm_{++}^d \to \reals$ is the `matrix fractional' function:
\[
h(y, X) = y^TX^{-1}y.
\]
Surprisingly, since the $\Sigma_i$ are positive semidefinite,
the corresponding equality constraint can also be written as a number of second order cone constraints, though this reduction
is slightly more complicated and we do not present it here. See, \eg,~\cite{lobo:1998} for more information.

\subsection{Safe quadrotor trajectory planning}
For higher order dynamical systems such as quadrotors, it is more desirable to plan entire safe trajectories rather that just finding a single safe point.  We can plan smooth polynomial trajectories for each agent $i$, such that the entire trajectory is inside each agent's reciprocally safe set, $P_i(t)$. After the polynomial trajectory is found, the required control inputs can be found via differential flatness~\cite{mellinger2011minimum}. In order to generate these trajectories and verify that they are safe, we require two extensions on the original CARP formulation: (1) a way to find a polynomial that is entirely inside the $P_i$, and (2) a method of expanding the uncertainty estimate $\ellipse^i_j$ to include the stopping set of other quadrotors.

\paragraph{B\'ezier curves.} Instead of the standard polynomial formulation, we use a B\'ezier curve to represent the polynomial trajectory. A $K$th order 3-dimensional B\'ezier curve, $B(t)\in \reals^3$, is defined by a set of $K+1$ ``control'' points $c_k \in \reals^d$, for $k=0,\dots,K$. We define the $K$th order B\'ezier curve as a linear combination of Bernstein polynomials:
\begin{equation}\label{eq:bezier}
    \begin{aligned}
    B(t) = \sum_{k=0}^K \binom{K}{k} (1-t)^{K-k} t^k c_k.
    \end{aligned}
\end{equation}
This curve has the property that it always lies in the convex hull of the control points, \ie, $B(t)\in \conv\{c_0,\dots, c_K\}$ for every $0 \le t \le 1$.
To find a polynomial of this form that is entirely inside the safe region, we can add additional constraints to~\eqref{eq:proj} to constrain each control point to be inside the set $P_i$. Finally, since the derivatives of a Bézier curve are linear combinations of the control points, we can account for the initial and desired final dynamic state (consisting of the position, velocity, and acceleration) by adding additional linear constraints to problem~\eqref{eq:proj}.

\paragraph{Stopping margins.} To include the stopping margins or each quadrotor, we simply expand the estimate $\ellipse^i_j$, similar to \cite{arul:2021}, by a sphere with radius $r=v_\mathrm{max}/2a_\mathrm{max}$, where $v_\mathrm{max}$ and $a_\mathrm{max}$ are the maximum velocity and acceleration of the quadrotor, respectively. This provides safe, yet conservative bound, for the stopping distance other agents will have during the planning horizon. In practice, if a velocity measurement is available then a more accurate inflation can be found. Since the original ellipsoidal estimate is only contains a point estimate of the other agents, we can expand the ellipsoid to account any number of arbitrary margins. Given a set of arbitrary ellipsoids $\ellipse_0,\dots,\ellipse_n$,  there exists an analytical method \cite{becis-aubry::2006,liu::2016} that finds the smallest (in the sense of major axes length) external bounding ellipsoid $\bar{\ellipse}$ for the Minkowski sum of all the ellipsoids $\bar{\ellipse} \supseteq \sum_{i=0}^n \ellipse_i$ . Thus we can, in a procedural and consistent manner, combine different margins.

\begin{figure}
    \centering
    \includegraphics[width=.44\textwidth]{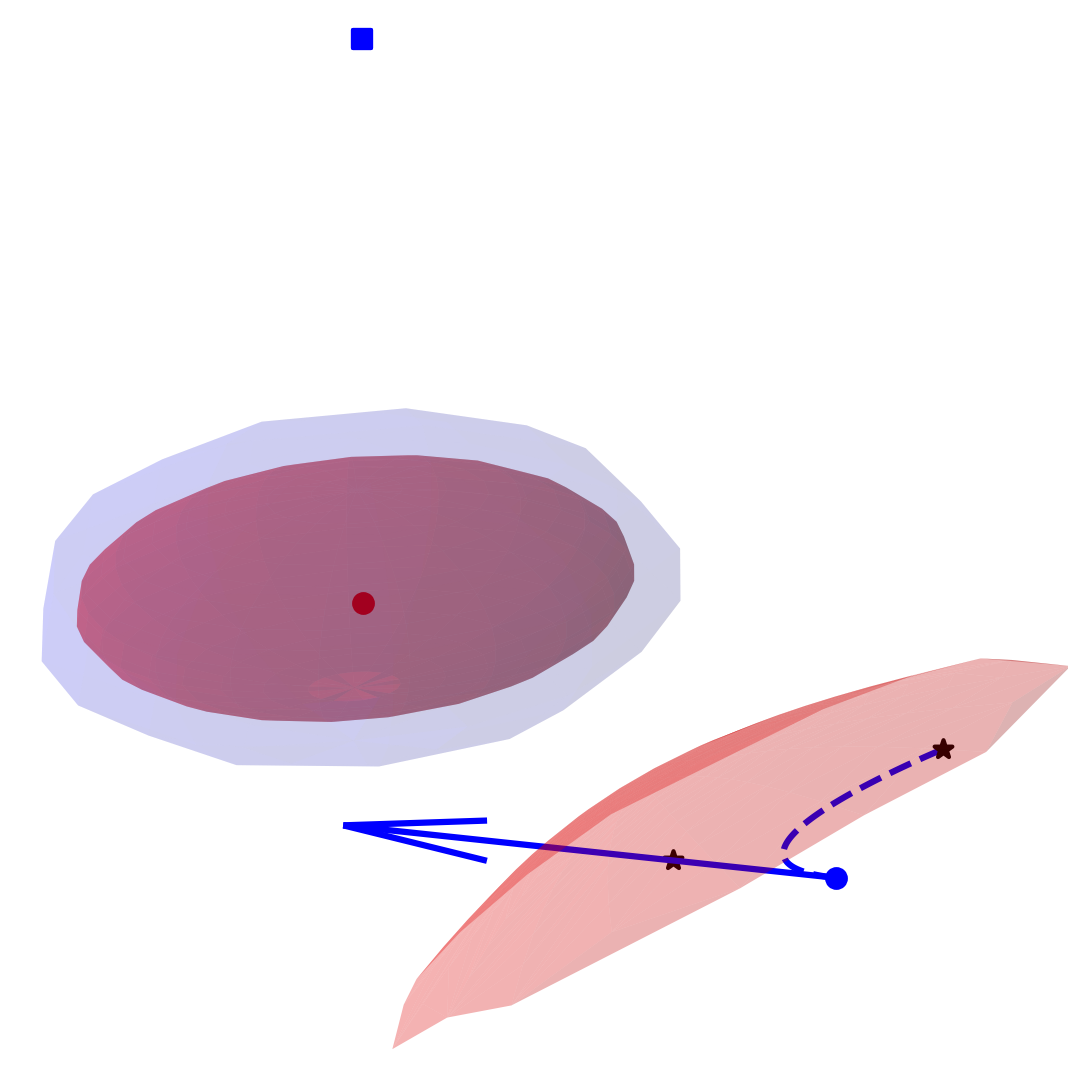} 
    \caption{Polynomial trajectory (blue, dotted) generated for an agent (blue circle) with non zero initial velocity (blue arrow) to avoid the ellipsoidal estimate of the obstacle (red ellipsoid) expanded by a margin (blue ellipsoid). The agent goal position (blue square) and the Bézier control points (black stars) are also shown.}
    \label{fig::projTraj}
\end{figure}

 \paragraph{Persistently safe receding horizon controller.} We extend problem~\eqref{eq:proj} to account for additional constraints placed on the $K$th order Bézier polynomial.
 To do this, we formulate the following optimization problem,
 \begin{equation}
    \label{eq:poly}
    \begin{aligned}
        & \text{minimize}
        & & \normsq{c_K-\xgoal} \\
        & \text{subject to}
        & & \normsq{x} - 2c_k^Tx + g(c_k, \lambda_k) + \lambda_k \le 0, \quad  k\in [K]\\
        & & & x = c_0 \\ 
        & & & v^0 = K (c_1-c_0) \\
        & & & v^f = K (c_K-c_{K-1}).
    \end{aligned}
\end{equation}
The optimization variables are $c_k \in \reals^3$ and $\lambda_k \ge 0$ for $k=0, \dots, K$, while $x$ is the current position, and
$v^0$ and $v^f$ are the current and desired final velocities, respectively. For compactness, we write $[K] = \{0, \dots, K\}$, and $g$ to be defined as in~\eqref{eq:ellipse}. As written, this finds a $1$ second long trajectory, but the section time for the trajectory can be changed by appropriately scaling the $t$ variable in~\eqref{eq:bezier}. Initial and final accelerations can also be added to~\eqref{eq:poly}. Since this
method is a refinement of~\eqref{eq:proj} (as any solution to~\eqref{eq:poly} is feasible for~\eqref{eq:proj} with $y=c_K$) then this method is guaranteed to be reciprocally safe.
% By definition of~\eqref{eq:r-safe} and the fact that the regions in a Voronoi tessellation are disjoint, the planning region the trajectories planned by each agent will be entirely in disjoint regions, eliminating the chance of collision. If the final velocity (and higher order dynamic states) are set to 0, then the paths are collision free for all time.
In the case where the optimization is infeasible, the agent can fall back to the previously calculated safe trajectory. Figure~\ref{fig::projTraj} shows an example instance of a polynomial trajectory being generated inside the safe-reachable area.

%results
\section{Results}\label{results}

\subsection{Projection Implementation}\label{sec:results:impl}
\begin{figure}
    \centering
    \includegraphics[width=0.6\textwidth]{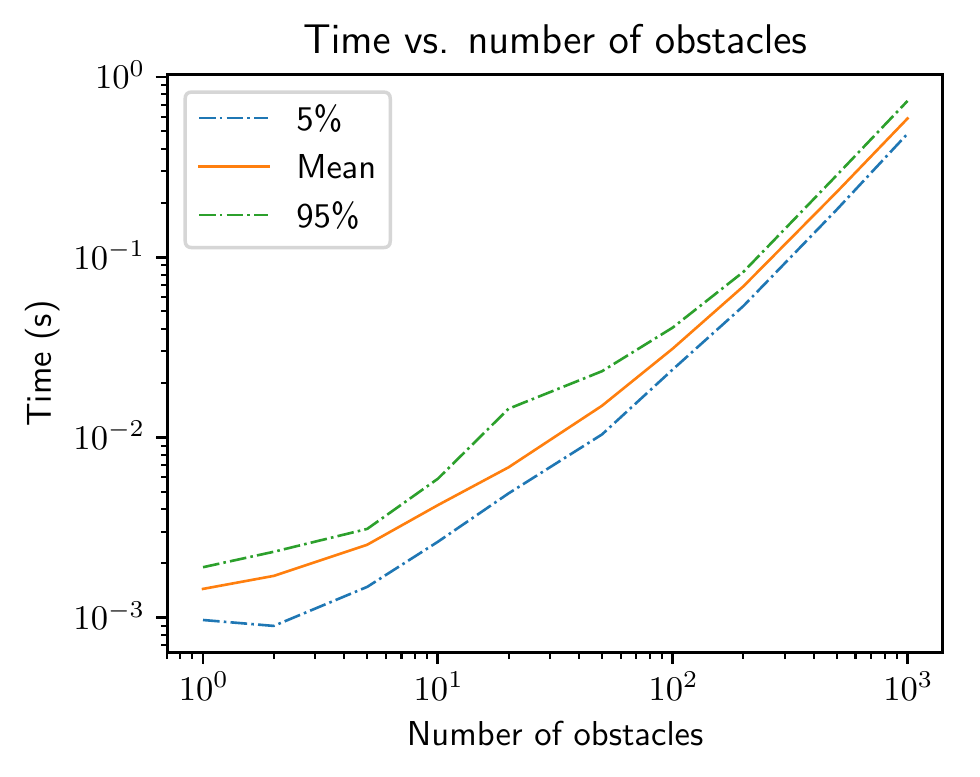}
    \captionof{figure}{Graph showing total time for generating and solving
    optimization problem~\eqref{eq:proj} as a function of the number of
    ellipsoids in the problem, when solved using the ECOS solver. Note the
    logarithmic scales on both axes.}
    \label{fig:timings}
\end{figure}

To get an accurate estimate of the speed of the projection algorithm,
the optimization problem outlined in~\eqref{eq:proj} was implemented\footnote{\url{https://github.com/angeris/CARP.jl}} in the
Julia language~\cite{bezanson:2017} using the JuMP mathematical programming
language~\cite{dunning:2017} and solved using ECOS~\cite{ecos:2013}.
We generated 285 instances of the problem, each with 100 randomly generated
ellipsoids in $\reals^3$. Timing and performance results for generating and
solving the corresponding convex program can be found in
table~\ref{tab:table_all}. Figure~\ref{fig:timings} shows how the performance
scales as the number of other agents increases. All times reported are on a
2.9GHz 2015 dual-core MacBook Pro.

\begin{center}
    \centering
    \begin{tabular}{@{}lr@{}}
        \textbf{Time} & \textbf{Total (GC \%)} \\
        Minimum & 13.20\si{\milli\second} (00.00\%)\\
        Median & 17.12\si{\milli\second} (00.00\%) \\
        Mean  & 17.55\si{\milli\second} (08.80\%) \\
        Maximum  & 36.77\si{\milli\second} (10.59\%) \\
    \end{tabular}
    \vspace{1em}
    \captionof{table}{Timings with garbage collection (GC) 
    as a percentage of time spent building and solving problems with 100 randomly 
    generated 3D ellipsoids. Statistics are based on
    285 instances and were obtained from the
    \texttt{BenchmarkTools.jl}~\cite{chen:2016} package.}
    \label{tab:table_all}
\end{center}

\subsection{Trajectory Simulations}
The projection algorithm was implemented in both 2D and 3D with a varying number
of agents. In this set up, each agent knows their own position exactly and
maintains a noisy estimate of other agents' positions, with uncertainties
represented as ellipsoids. This estimate is updated by a set-membership based
filter~\cite{bertsekas::1971,liu::2016,shah:2019}, a variant of the Kalman
filter. We expand the uncertainty ellipsoid by a given margin to account for the
robot's physical size.  If this margin is also ellipsoidal, then a small
ellipsoid which contains the Minkowski sum of the uncertainty ellipsoid and the
margin can be found in closed form~\cite{liu::2016,shah:2019}. This new bounding
ellipsoid is used in the projection algorithm to account for a user defined
margin, along with the uncertainty ellipsoid containing the noisy sensor
information.

\begin{figure}
	\centering
    \includegraphics[width=.48\textwidth]{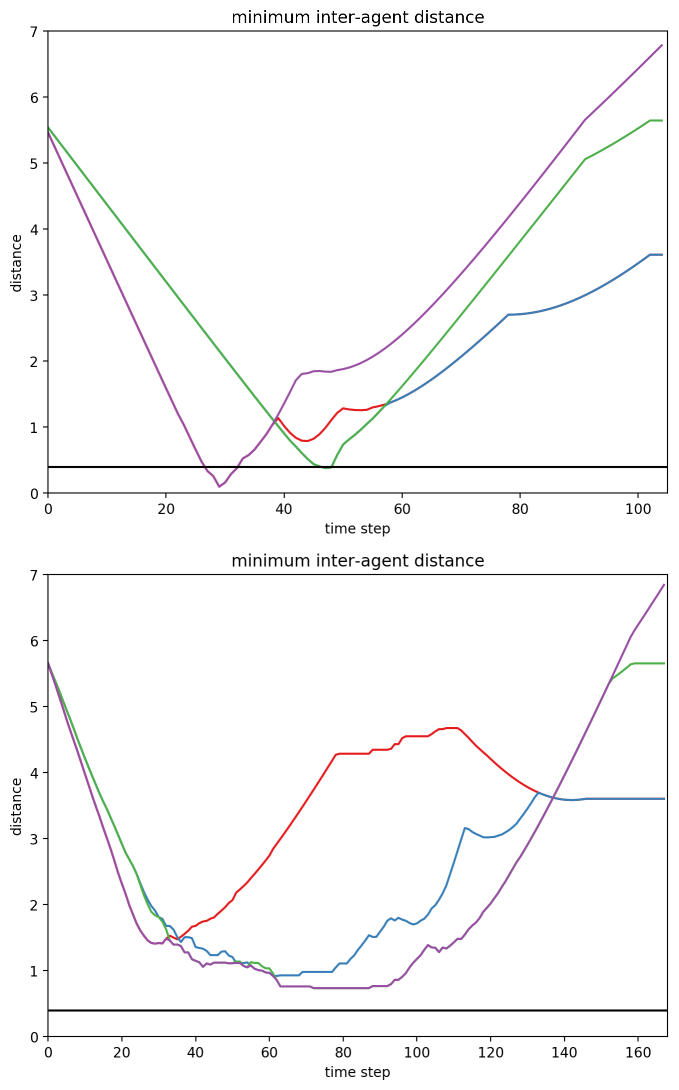}
	\caption{Inter-agent distances RVO (top) and our method (bottom). The RVO 
	simulation results in inter-agent distances below the collision threshold 
	(black line).}
	\label{fig::2dsim-dist}
\end{figure}

Figure~\ref{fig::2dsim-dist} shows the minimum inter-agent distances for each 
agent in the simulation scenario mentioned above. The collision threshold was 
set to $.4\si{\meter}$, twice the radius of the agents. Although our method 
results in longer paths, it remains collision free, while RVO's paths result in collision.

\begin{figure*}
\centering
    \begin{minipage}[]{0.8\textwidth}
    \centering
    \includegraphics[width=.45\textwidth]{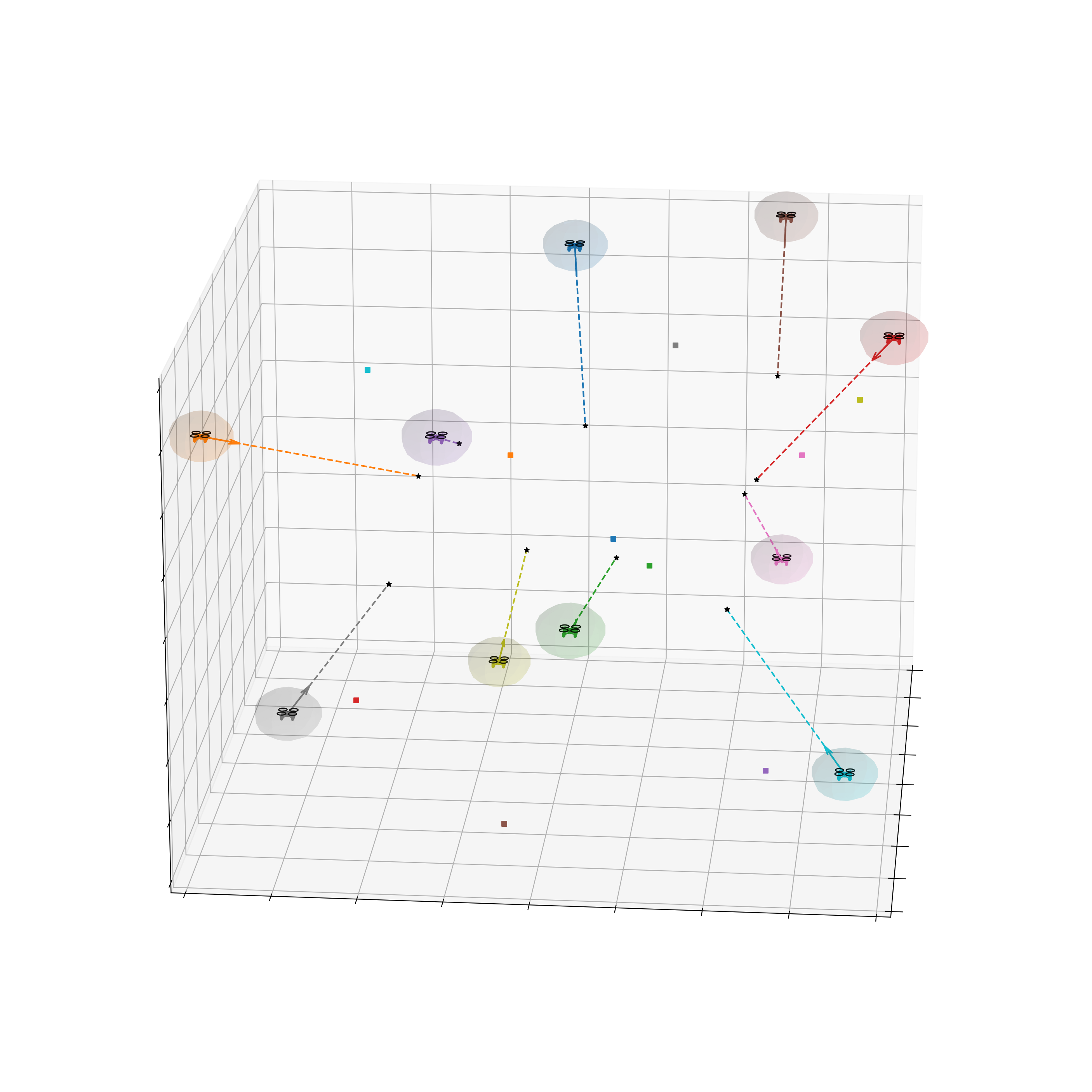}
    \includegraphics[width=.45\textwidth]{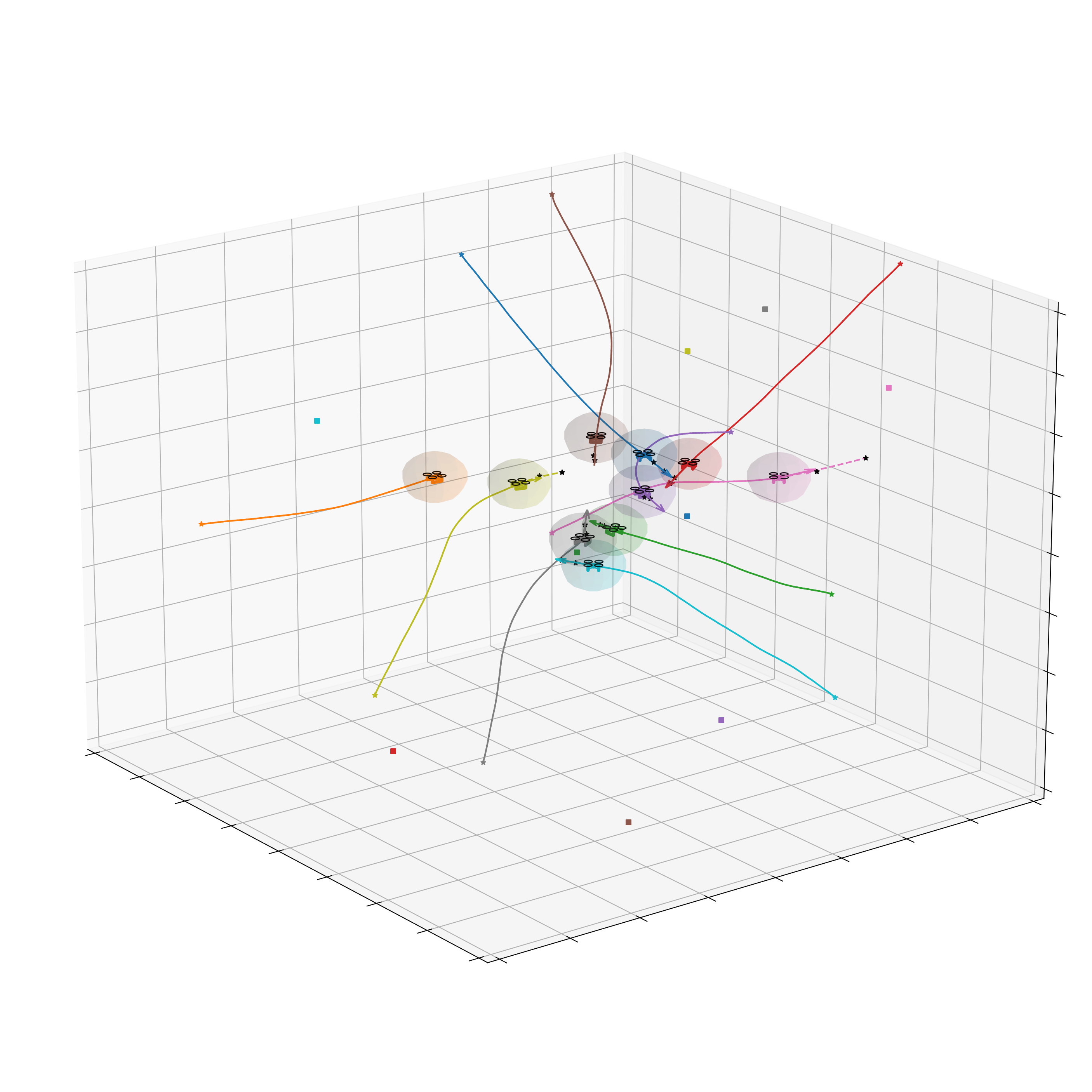}
    $t=0\hspace{2in} t=60$\\
    \includegraphics[width=.45\textwidth]{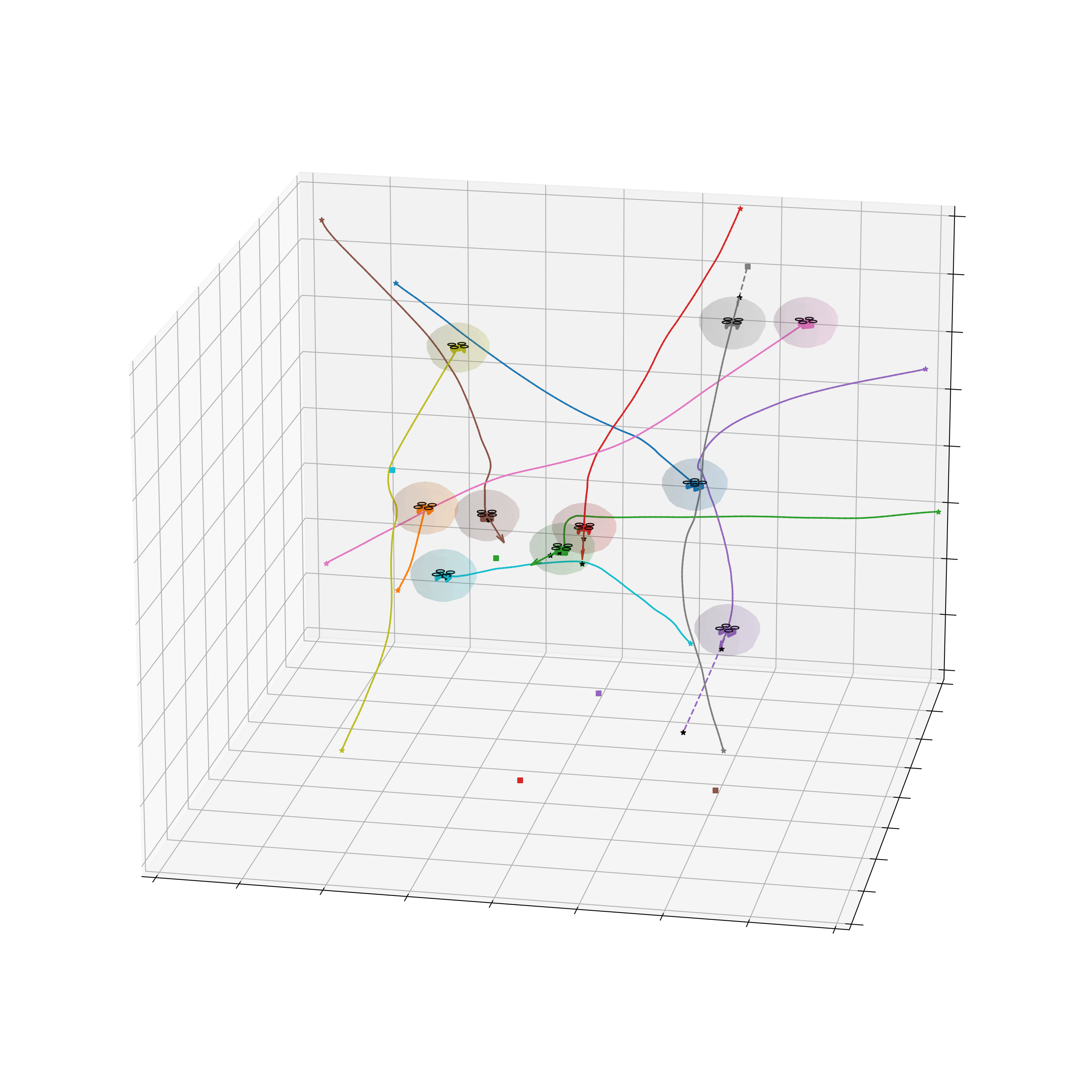}
    \includegraphics[width=.45\textwidth]{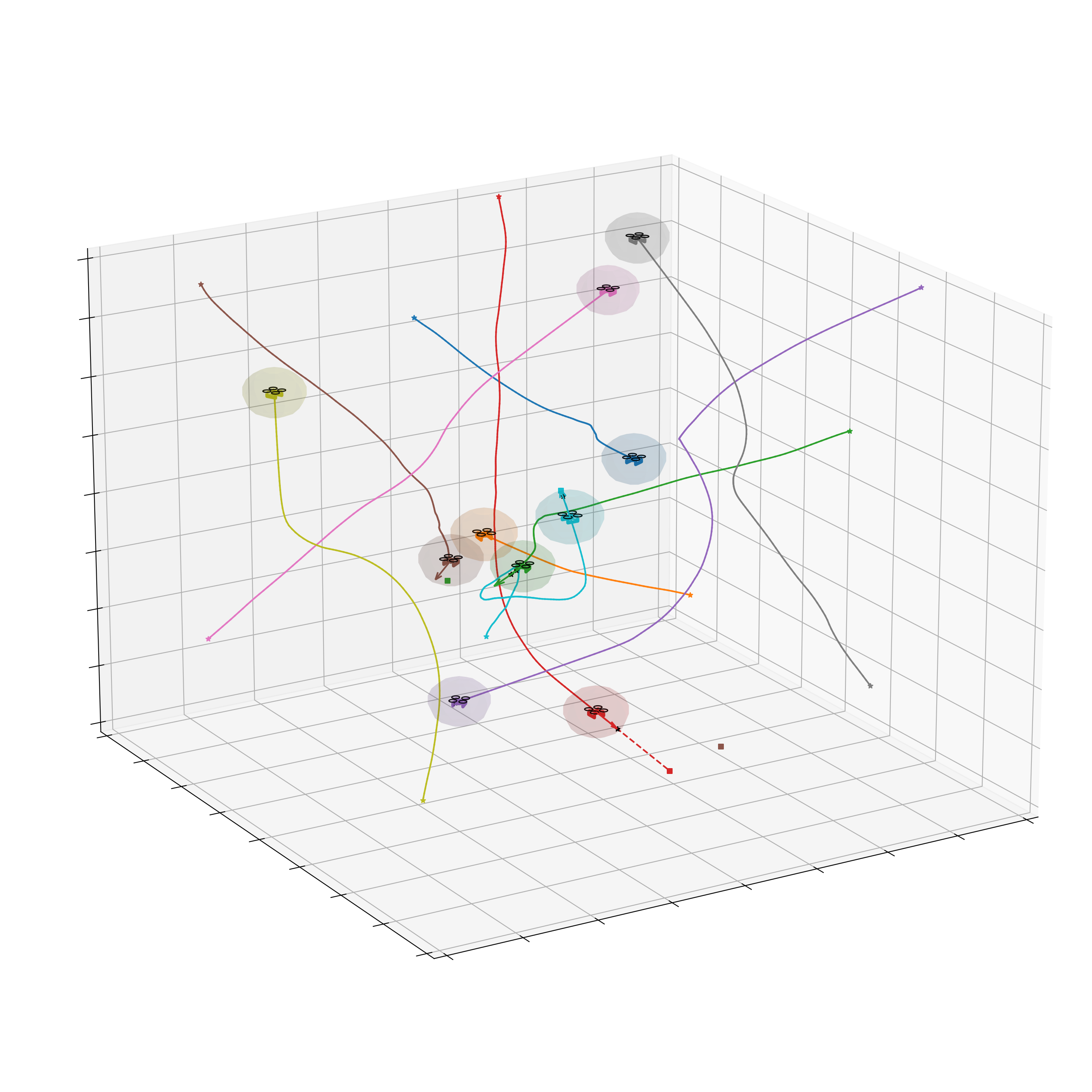}
    $t=120\hspace{2in} t=180$\\
    \includegraphics[width=.45\textwidth]{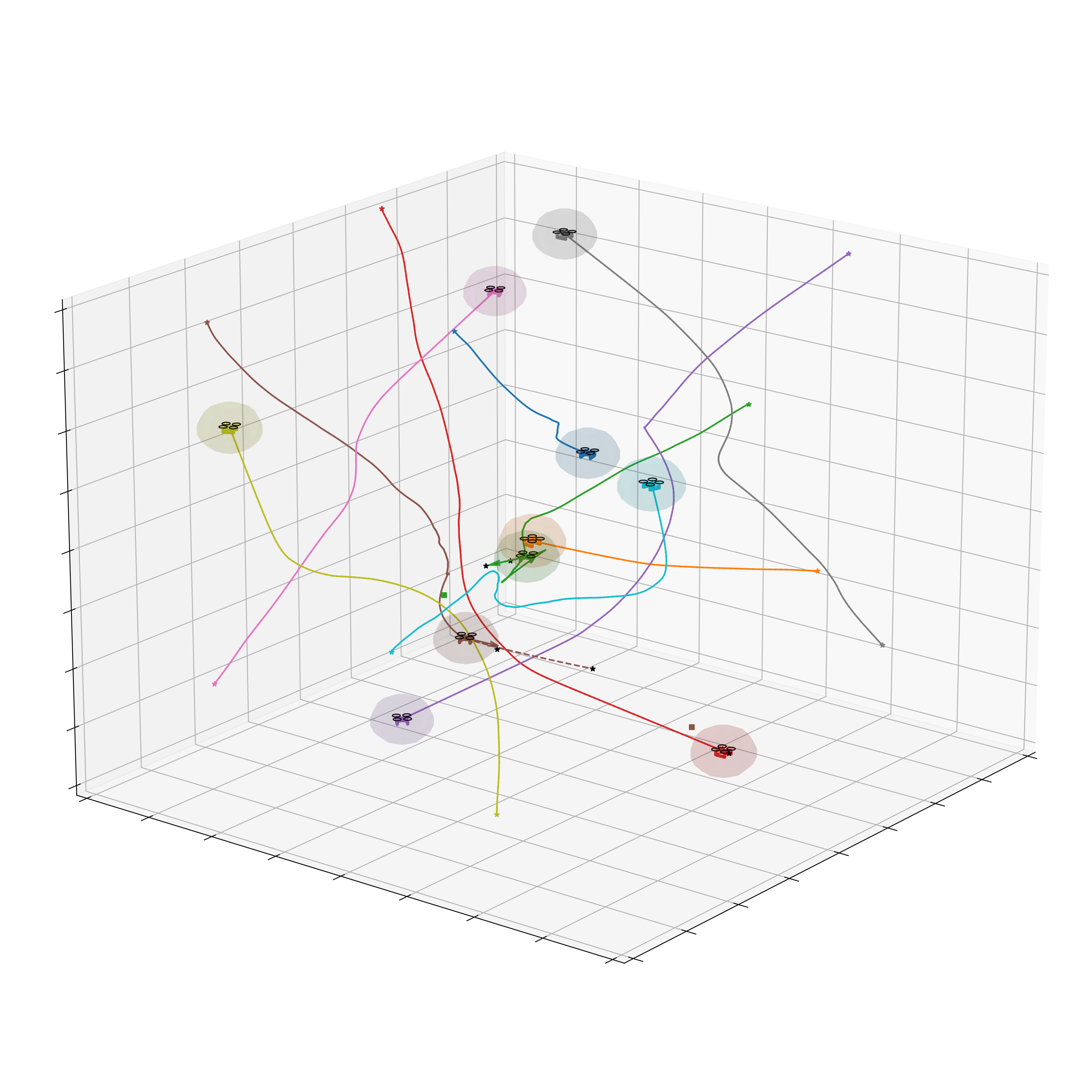}
    \includegraphics[width=.45\textwidth]{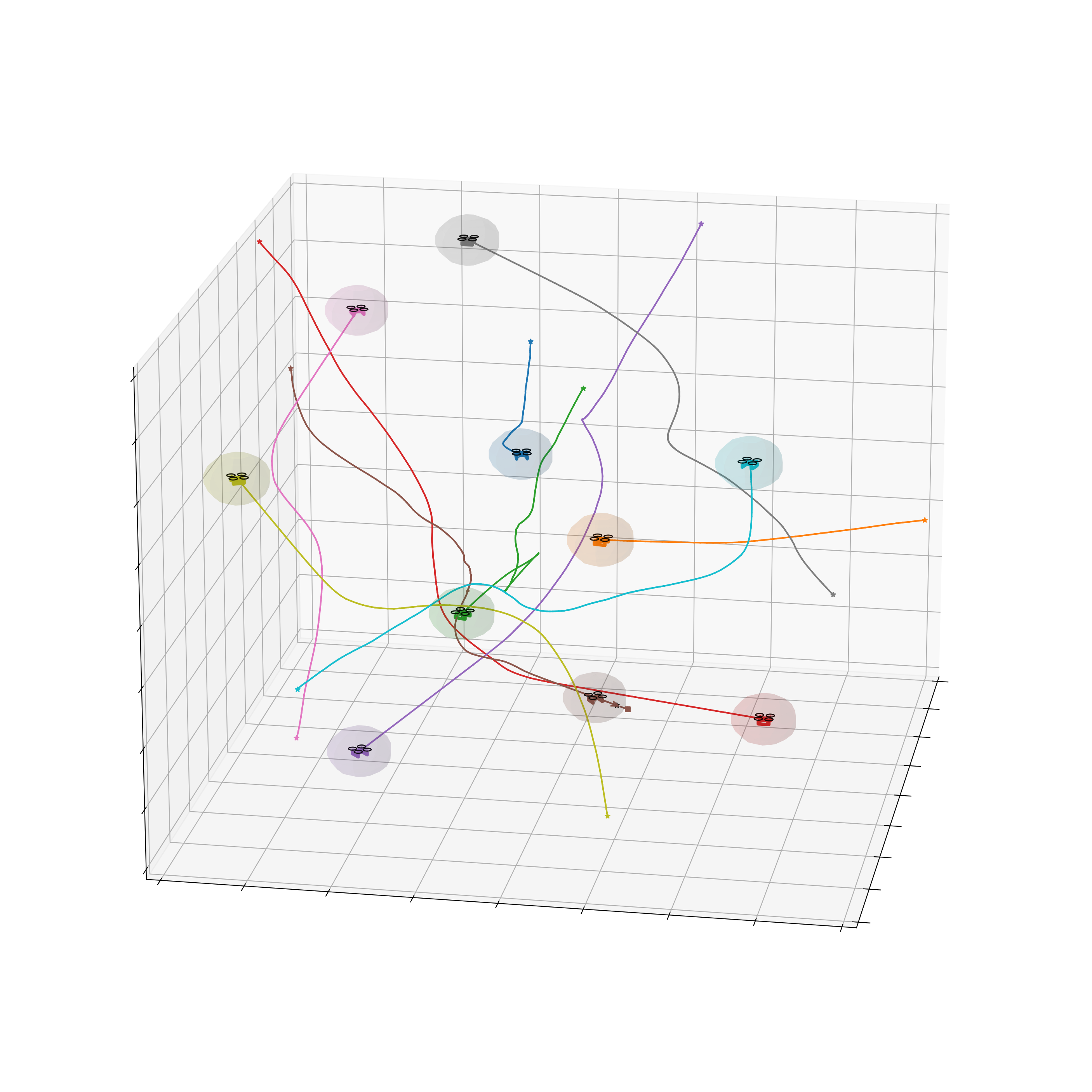}
    $t=210\hspace{2in} t=250$
    \vspace{.2in}
    \caption{Six time instances of a 3D simulation of 10 agents. Each agent 
    adds an ellipsoidal margin (shown) elongated in the z-axis to account for 
    downwash affects.}
    \label{fig::3dsim}
    \end{minipage}
\end{figure*}

Figure~\ref{fig::3dsim} shows six instances of a 3D simulation with 10
agents using the polynomial method. The agents start at the sides of a
$10\si{\meter}\times10\si{\meter}\times10\si{\meter}$ cube and are constrained
to a maximum speed of $6\si{\meter/\second}$ and a maximum measurement error set
to $1.0\si{\meter}$.The agents, displayed as quadrotors, each have a bounding box of $0.45\si{\meter}\times0.45\si{\meter}\times0.2\si{\meter}$ and an additional ellipsoidal margin with an axis length of $.3\si{\meter}$ in the $x$and $y$ dimensions, and $.7\si{\meter}$ in the $z$ dimension.
This margin effectively gives a buffer of $0.75\si{\meter}$ in the $xy$ plane
and a large buffer of $.9\si{\meter}$ in $z$. We assume
non-spherical margins in this simulation since, in the case of quadrotor flight,
large margins in the $z$ direction can prevent unwanted effects due to
downwash~(\cite{downwash:2017}). The minimum inter-agent distance during the
simulation, as measured from the centers of the agents, was $1.6\si{\meter}$.

\subsection{Hardware Demonstrations}

\begin{figure}[H]
    \centering
    \includegraphics[width=.48\textwidth]{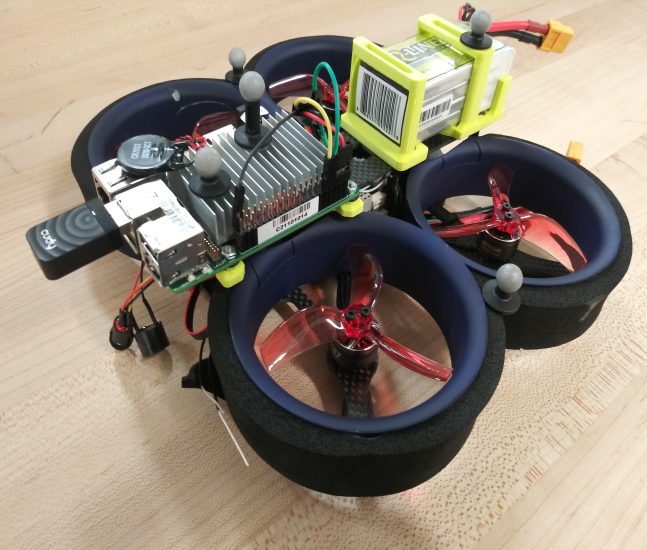}
    \caption{Quadrotor equipped with a PX racer and UP Board companion computer.}
    \label{fig::quad}
\end{figure}

\begin{figure}[htb]
    \centering
    \includegraphics[width=.48\textwidth]{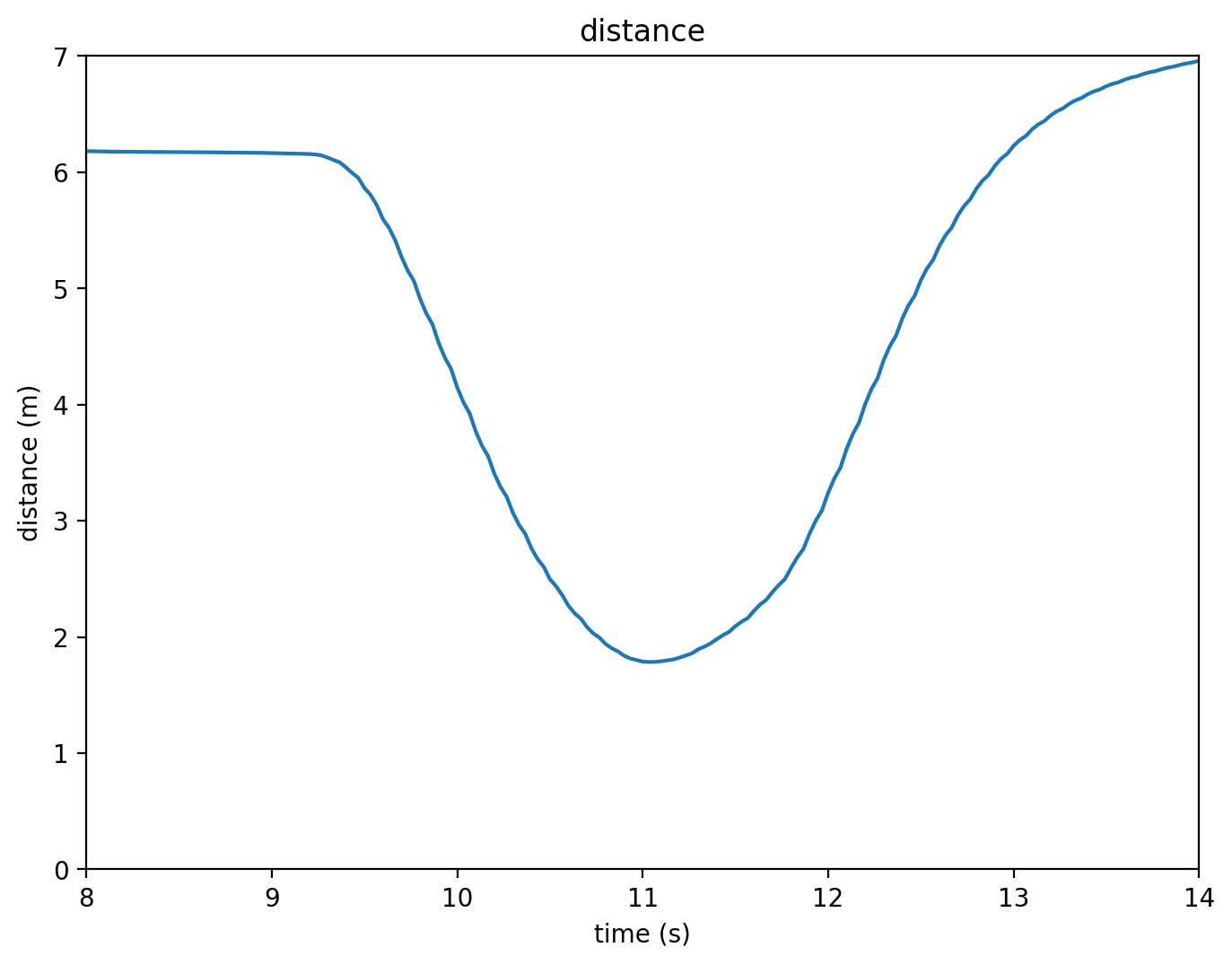}
    \\
    \includegraphics[width=.48\textwidth]{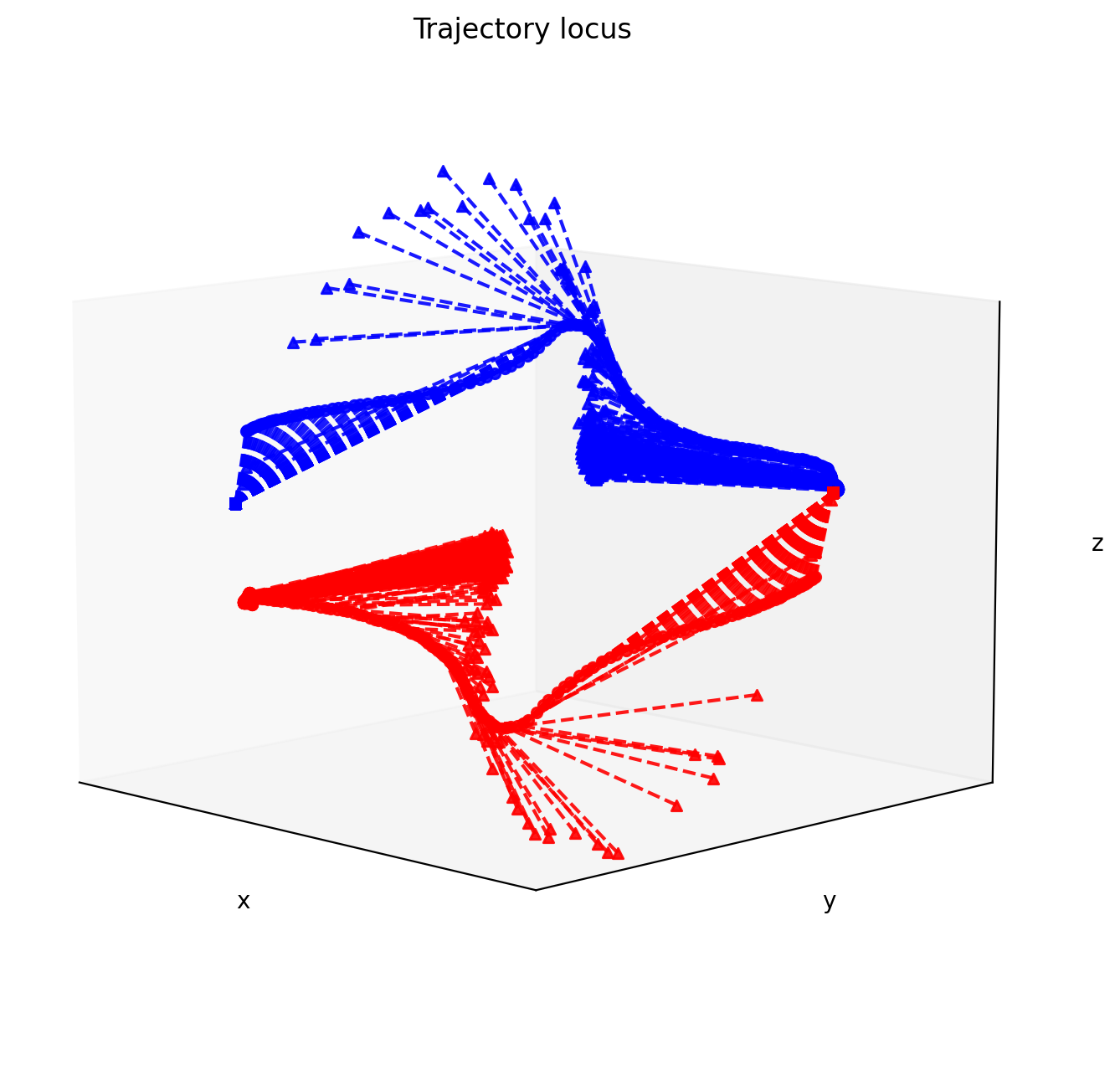}
    \caption{Projection only method: Inter-agent distance (top) and trajectory locus (bottom) for two quadrotors. The minimum distance between agents is $1.78\si{\meter}$. The locus plot show the reprojected point (triangle) and goal point (square) over the course of the flight.}
    \label{fig::proj_result}
\end{figure}

\begin{figure}[htb]
    \centering
    \includegraphics[width=.48\textwidth]{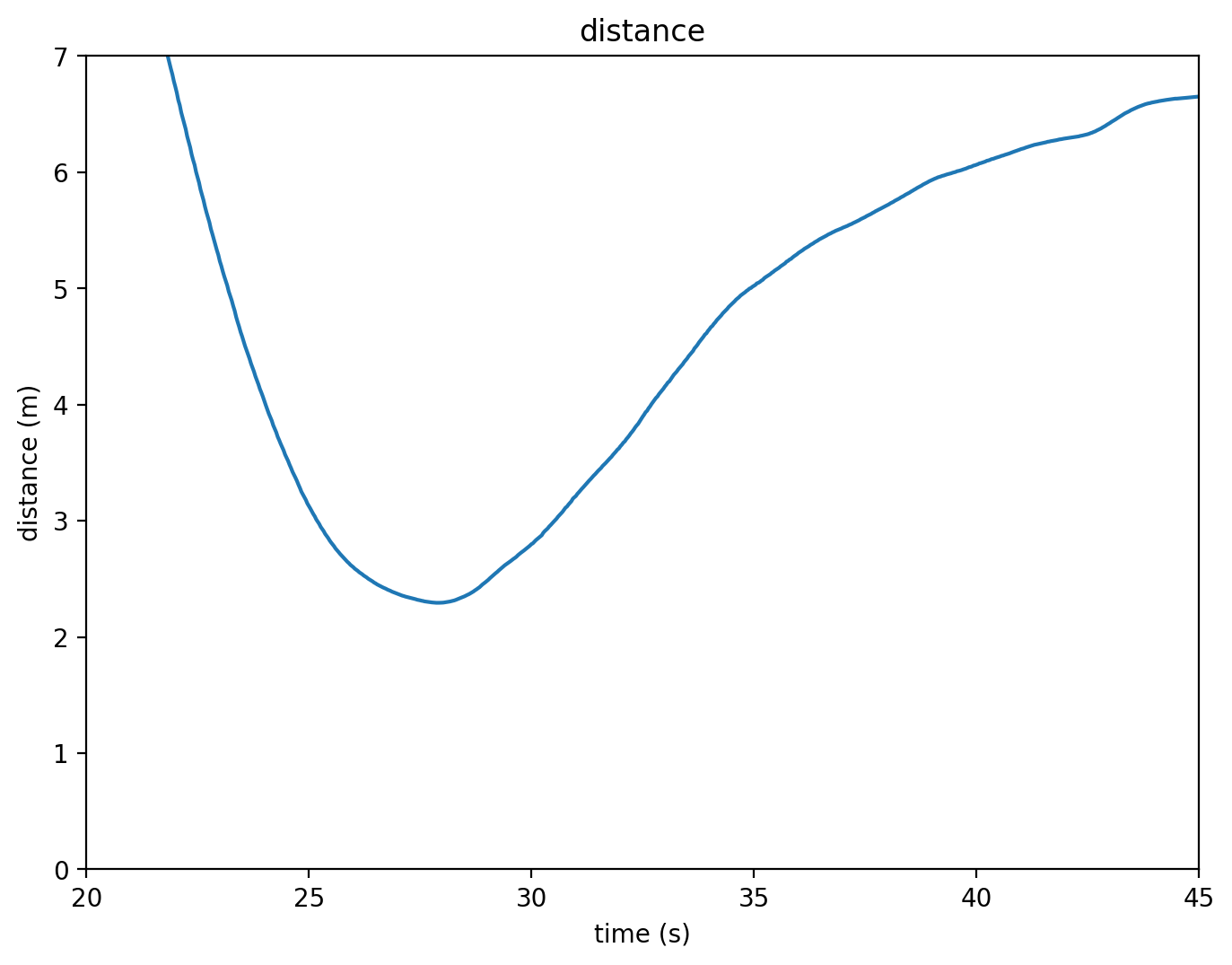}
    \\
    \includegraphics[width=.48\textwidth]{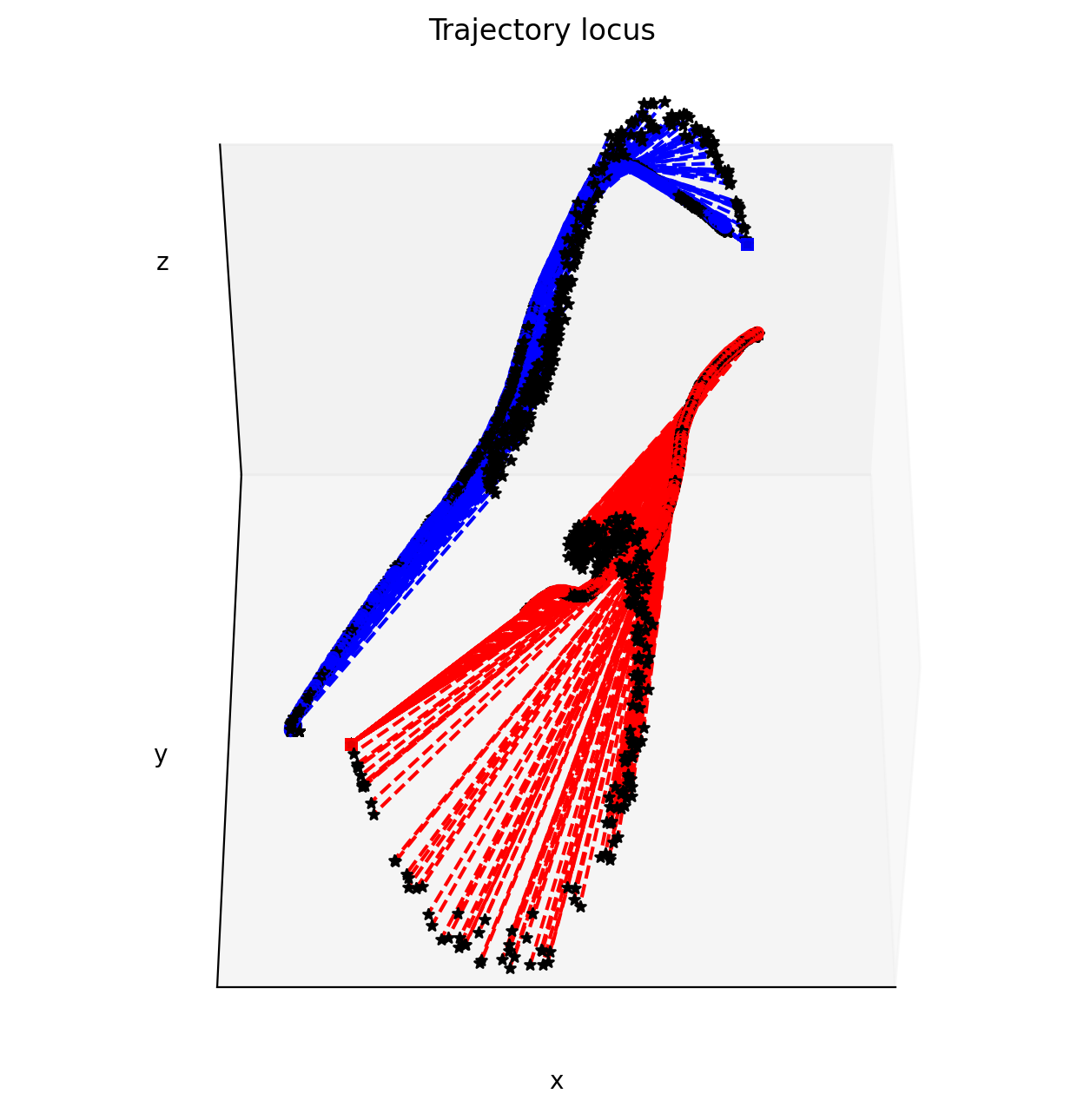}
    \caption{B\'ezier polynomial method: Inter-agent distance (top) and trajectory locus (bottom) for two quadrotors. The minimum distance between agents is $2.29\si{\meter}$. The locus plot show the trajectory (dotted) and control points (star) as well as the final goal point (square) over the course of the flight.}
    \label{fig::poly_result}
\end{figure}

We implemented\footnote{\url{https://github.com/StanfordMSL/carp_ros}} our collision avoidance the method on a set of 2 PX4 based quadrotor platforms (figure \ref{fig::hero}) to demonstate our method on a real robotic system\footnote{A video of the simulations and experiments can be found at
\url{https://youtu.be/oz-bMovG4ow}.}. 
Each quad (see figure~\ref{fig::quad}) is outfitted with a x86 based UP board and is tracked via a motion capture system. Each agent is sent its own position 
and maintains a noisy ellipsoidal estimate other agents' positions, which are filtered by a set-membership filter~\cite{bertsekas::1971,liu::2016,shah:2019} and 
updated via noise-corrupted measurements. Planning and closed loop trajectory control, as well as estimation and filtering, are all done on-board.
The planner is 
updated at $80\si{hz}$, while the filter is run at $100\si{hz}$. Low-level control is done with a feed-forward PID velocity controller. We implemented both the 
direct goal projection method as well as the polynomial trajectory generation method. The quadrotors measure $23\si{\cm}$ by $23\si{\cm}$ by $2.3\si{\cm}$ and 
are given an additional $30\si{\cm}$ margin in the $xy$ plane and a $60\si{\cm}$ margin in the $z$ axis. Figures~\ref{fig::proj_result} 
and~\ref{fig::poly_result} show the distance between agents over the course of the flight, as well as the final trajectory and the intermediate trajectories for 
the projection method (figure~\ref{fig::proj_result}) and the polynomial method (figure~\ref{fig::poly_result}). 

%conc
\section{Closing Remarks}
%While cloud computing has relieved many burdens distributed robotic
%systems encounter, network delays, disconnects,
%and other failures are still commonplace at scale. Dependable and fast
%algorithms that run on-board are therefore %critical for any
%certifiable system. 
In this work, we presented a scalable system that
can work with simple or complex, distributed or centralized high-level
planners to provide safe trajectories for a team of agents. Under the
assumptions stated, we showed that collision avoidance is guaranteed
provided each agent follows this method. However, we observe practical
collision avoidance behavior even if only the ego agent follows this
method. Computational performance results and simulations provide
evidence that this algorithm can potentially be used in
safety-critical applications for mobile robots with simple dynamics. Our open source
library can also be used directly as a ROS package.

%Future work will focus on analyzing the complex feedback between the control 
%and estimator. We also plan to create a fast, easily extendable library for
%automatically generating programs of the form of problem~\eqref{eq:proj}, by
%making use of the composition rules presented in the appendix. Finally, by incorporating 
%global path-planners---which handle static obstacles less conservatively than our 
%algorithm---directly into our trajectory-planning pipeline, we aim to have this 
%library handle the motion planning for a variety of different dynamic agents in 
%various environments.
%We suspect that
%future support for warm starts and the ability to change parameters without
%reconstructing the complete problem from scratch (as compared
%to~\S\ref{sec:results:impl}) would yield a substantial speed up in solution
%time. 
%Additionally, we note that while immediately attempting to use
%problem~\eqref{eq:final} in a Model Predictive Control (MPC) formalism yields a
%nonconvex problem, we believe that there may be a approach to approximate
%solution while retaining the feasibility properties assumed
%in~\eqref{eq:proj}. 
%We also plan on incorporating a global planner that
%handles static obstacles less conservatively than our algorithm into our
%trajectory-planning pipeline.

\textbf{Acknowledgments.}
This work was supported in part by the Ford-Stanford Alliance program, and by DARPA YFA award D18AP00064. Guillermo Angeris is supported by the National Science Foundation Graduate Research Fellowship under Grant No.\ DGE-1656518.
% ---- Bibliography ----
\clearpage
\bibliographystyle{alpha}
\bibliography{cites}
%appendix
% \newpage
% \input{sections/7_appendix.tex}
\end{document}